\makeatletter\def\graphicscache@inhibit{true}\makeatother
\documentclass{article}
\usepackage[utf8]{inputenc}

\usepackage{jfrExamplee}
\usepackage{graphicx}
\usepackage{natbib}
\let\bibhang\relax
\usepackage{apalike}
\usepackage{setspace}

\usepackage{array}

\usepackage[hidelinks]{hyperref}
\usepackage[capitalise]{cleveref}
\usepackage[draft]{fixme}
\fxsetup{theme=color}
\usepackage{booktabs}
\usepackage{amsmath}
\usepackage{placeins}
\newcommand\w[1]{\mathbf{#1}}

\usepackage{color}
\usepackage[table]{xcolor}
\usepackage{amssymb}
\usepackage{siunitx}
\usepackage{balance}
\usepackage{paralist}
\usepackage{xspace}
\usepackage{multirow}
\usepackage{tabularx}
\usepackage{subfig}
\usepackage{tabu}
\usepackage{threeparttable}

\newcommand{\eg}{e.g.,\ }
\newcommand{\ie}{i.e.\ }
\newcommand{\reffig}[1]{Fig.~\ref{#1}}

\usepackage{tikz}
\usetikzlibrary{arrows}
\usetikzlibrary{positioning,calc}
\usetikzlibrary{decorations.pathreplacing}
\usetikzlibrary{decorations.markings}
\usetikzlibrary{fit}
\usetikzlibrary{shapes.callouts}

\tikzset{
  big arrow/.style={
    decoration={markings,mark=at position 1 with {\arrow[scale=1.7]{latex}}},
    postaction={decorate},
    shorten >=0.4pt}}
\tikzset{
  big 2arrow/.style={
    decoration={markings,mark=at position 0 with {\arrow[scale=-1.7]{latex}},mark=at position 1 with {\arrow[scale=1.7]{latex}}},
    postaction={decorate},
    shorten >=0.4pt}}

\usepackage{pgfplots}
\pgfplotsset{compat=1.9}
\usepackage{csvsimple}

\usepackage{enumitem}

\usepackage{tablefootnote}

\pgfdeclarelayer{background}
\pgfdeclarelayer{foreground}
\pgfsetlayers{background,main,foreground}

\usepackage[tightpage]{preview}
\newenvironment{maybepreview}%
{\noindent\ignorespaces}%
{\par\noindent%
\ignorespacesafterend}

\usepackage[multidot]{grffile}

\usepackage[qfactor=0.4]{graphicscache}

\title{NimbRo Rescue: Solving Disaster-Response Tasks\\ through Mobile Manipulation Robot Momaro}

\author{
Max Schwarz, Tobias Rodehutskors, David Droeschel, Marius Beul, Michael Schreiber,  \\
\textbf{Nikita Araslanov, Ivan Ivanov, Christian Lenz, Jan Razlaw, Sebastian Sch{\"u}ller,} \\
\textbf{David Schwarz, Angeliki Topalidou-Kyniazopoulou, and Sven Behnke} \\
Autonomous Intelligent Systems Group \\
University of Bonn \\
Bonn, Germany \\
\texttt{max.schwarz@uni-bonn.de}, \texttt{\{firstname.lastname\}@ais.uni-bonn.de} \\
}

\begin{document}

\maketitle

\begin{tikzpicture}[remember picture,overlay]
\node[anchor=north west,align=left,font=\sffamily,yshift=-0.2cm] at (current page.north west) {%
  Journal of Field Robotics (JFR) 34(2):400–425, 2017
};
\node[anchor=north east, align=right,font=\sffamily,yshift=-0.2cm] at (current page.north east) {%
  DOI: \href{https://doi.org/10.1002/rob.21677}{10.1002/rob.21677}
};
\end{tikzpicture}%

\begin{abstract}
Robots that solve complex tasks in environments too dangerous for humans to 
enter are desperately needed, \eg for search and rescue applications.
We describe our mobile manipulation robot Momaro, with which we participated
successfully in the DARPA Robotics Challenge. It features a unique
locomotion design with four legs ending in steerable wheels, which allows
it both to drive omnidirectionally and to step over obstacles or climb.
Furthermore, we present advanced communication and teleoperation approaches,
which include immersive 3D visualization, and 6D tracking of operator head and
arm motions.
The proposed system is evaluated in the DARPA Robotics Challenge, the DLR
SpaceBot Cup Qualification and lab experiments. We also discuss the lessons
learned from the competitions.
\end{abstract}

\section{Introduction}

Disaster scenarios like the Fukushima nuclear accident clearly reveal the need 
for robots that are capable to meet the requirements arising during
operation in real-world, highly unstructured and unpredictable situations, where 
human workers cannot be deployed due to radiation, danger of collapse or toxic 
contamination. As a consequence of the incident in Fukushima, the Defense 
Advanced Research Projects Agency~(DARPA) held the DARPA Robotics 
Challenge\footnote{\url{http://web.archive.org/web/20160420224150/http://theroboticschallenge.org/} (original unavailable)}~(DRC) to foster the 
development of robots capable of solving tasks which are required to relief 
catastrophic situations and to benchmark these robots in a competition. During 
the DRC, the robots needed to tackle eight tasks within one hour: 1. Drive a 
vehicle to the disaster site, 2. Egress from the vehicle, 3. Open a door, 4. 
Turn a valve, 5. Cut a hole into a piece of drywall, 6. Solve a surprise 
manipulation task, 7. Overcome rough terrain or a field of debris, and 8. Climb 
some stairs. To
address this large variety of tasks, we constructed the mobile manipulation robot Momaro
and an accompanying teleoperation station for it.

\begin{figure}[t]
  \centering
  \begin{maybepreview}%
\begin{tikzpicture}
    \node[anchor=south west,inner sep=0] (image) at (0,0) {\includegraphics[height=10cm]{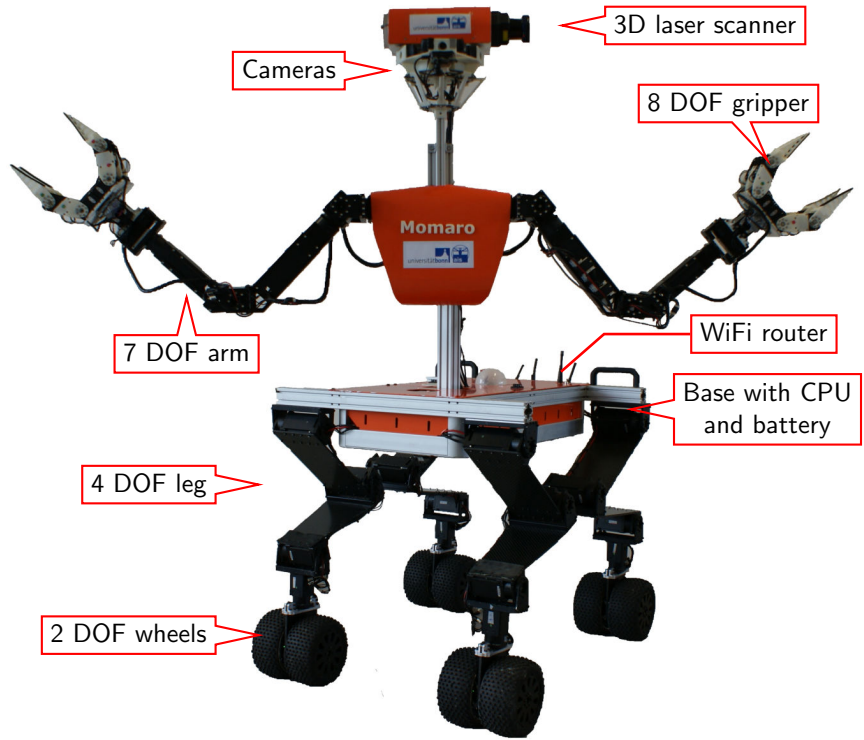}};
    \begin{scope}[
            x={(image.south east)},
            y={(image.north west)},
            font=\sffamily,
            every node/.style={align=center},
            line/.style={red, line width=1pt},
            box/.style={rectangle,draw=red,inner sep=0.3333em,thick},
every node/.style={align=center,text height=1.5ex,
    text depth=.25ex,
    text centered
    },
          ]
        \node[box,anchor=west,rectangle callout,callout relative pointer={(-0.05,0.0)}] (cam) at (0.7, 0.96) {3D laser scanner};
        \node[box,anchor=east,rectangle callout,callout relative pointer={(0.05,0.0)}] (cam) at (0.4, 0.9) {Cameras};
        \node[box,anchor=east,rectangle callout,callout relative pointer={(0.05,0.0)}] (cam) at (0.25, 0.35) {4 DOF leg};
        \node[box,anchor=east,rectangle callout,callout relative pointer={(0.05,0.0)}] (cam) at (0.25, 0.15) {2 DOF wheels};
        \node[box,anchor=west,rectangle callout,callout relative pointer={(-0.05,0.0)},text height=] (cam) at (0.78, 0.45) {Base with CPU\\and battery};

        \node[box,anchor=north east,rectangle callout,callout relative pointer={(0.0,0.05)}] at (0.3,0.55) {7 DOF arm};

        \node[box,anchor=south east,rectangle callout,callout relative pointer={(0.03,-0.05)}] at (0.95,0.83) {8 DOF gripper};

        \node[box,anchor=west] (wifi) at (0.8, 0.55) {WiFi router};
        \draw[line] (wifi) -- (0.7,0.55) -- (0.65,0.5);
    \end{scope}
\end{tikzpicture}
\end{maybepreview}
  \caption{The mobile manipulation robot Momaro.}
  \label{fig:momaro}
\end{figure}

Momaro (see \cref{fig:momaro}) is equipped with four articulated compliant legs
that end in pairs of directly driven, steerable wheels. 
This unique base design combines advantages of driving and stepping locomotion.
Wheeled systems, which include also tank-like tracked vehicles, are robust and
facilitate fast planning, while being limited in the height differences or terrain types they can overcome.
Legged systems require more effort to control and maintain stability, but
can cope with quite difficult terrain, because they require only isolated safe
footholds. On the downside, they often move slower than wheeled systems.
Hybrid systems with a combination of legs and wheels, namely legs ending in
wheels, promise to combine the benefits of both locomotion modes. On sufficiently smooth terrain,
locomotion is done by driving omnidirectionally on the wheels while adapting to
slow terrain height changes with the legs. If larger obstacles prevent driving,
the robot switches to stepping locomotion. With these advantages in mind,
we chose a hybrid locomotion scheme for Momaro.

To perform a wide range of manipulation tasks, Momaro has an anthropomorphic
upper body with two 7~degrees of freedom~(DOF) manipulators that end in dexterous grippers.
This allows for the single-handed manipulation of smaller objects, as well as for two-armed manipulation of larger objects and the use of tools.
Through adjustable base height and attitude and a yaw joint in the spine, Momaro has a work space equal to the one of an adult person.

The DRC requirements are beyond the state of the art of autonomous robotics. As 
fully autonomous systems which work in these complex environments are not 
feasible yet, often human intelligence is embedded into the robot through 
teleoperation to improve the overall performance of the system. Human operators 
can easily react to unforeseen events, but require awareness of the situation. 
To this end, we equipped our robot with a 3D laser scanner, multiple 
cameras, a microphone for auditory feedback, and an infrared distance sensor.

For effective teleoperation of the many DOFs of our robot, intuitive and flexible user interfaces are key.
For driving the car, multiple cameras and the visualization of the 3D scene provide good 
situation awareness and the operator can control the car directly using a steering wheel and a gas pedal.
The motions of these remote controllers are mapped to robot limbs actuating the corresponding car controllers.
Omnidirectional driving is controlled using a three-axis joystick, based on camera and 3D scene feedback.
The velocity commands are mapped to the directions and speeds of the eight robot wheels. 
To solve complex bimanual manipulation tasks, we developed a teleoperation interface consisting of a stereoscopic head-mounted display (HMD) 
and two 6D magnetic trackers for the hands of the operator. The operator head 
motions are tracked to render views based on the available 3D point clouds for 
the HMD, which follow his motions with low latency and therefore increase his feeling of
immersion. The position and orientation of the magnetic trackers are mapped to 
the end-effectors of our robot using inverse kinematics with redundancy 
resolution to calculate positional control commands for Momaro's anthropomorphic 
arms. 
For the indoor tasks 4-7, DARPA degraded the communication between the
operators and the robot, and data transmission had to be carefully managed. 
To address this communication restriction, we developed a method for combining a low-latency low-bandwidth 
channel with a high-latency high-bandwidth channel to provide the operators high-quality low-latency situation awareness. 

All the developed components were integrated to a complete disaster-response system, which performed very well at the DARPA Robotics Challenge. Through Momaro, our team NimbRo Rescue
solved seven of the eight DRC tasks in only 34 minutes, coming in as best
European team at the 4th place overall.
We report in detail on how the tasks were solved.
The system was also tested in the DLR SpaceBot Cup and in lab experiments.
Our DRC developments led to multiple contributions, which are summarized in this article, including the unique hybrid locomotion concept, good situation awareness despite degraded communication, and intuitive teleoperation interfaces for solving complex locomotion and manipulation tasks.
We also discuss lessons learned from the challenging robot operations.

\section{Related Work}
\label{sec:related}

The need of mobile manipulation has been addressed in the past with the
development of a variety of mobile manipulation systems consisting of robotic arms
installed on mobile bases with the mobility provided by wheels, tracks, or leg
mechanisms. Several research projects exist which use purely wheeled locomotion
for their robots~\citep{Mehling07,Borst09}. In previous work, we developed
NimbRo Explorer~\citep{stuckler2015nimbro}, a six-wheeled robot equipped with a
7\,DOF arm designed for mobile manipulation in rough terrain encountered in
planetary exploration.

Wheeled rovers provide optimal solutions for well-structured, and relatively
flat environments, however, outside of these types of terrains,
their mobility quickly reaches its limits. Often they can only overcome
obstacles smaller than the size of their wheels. Compared to wheeled robots,
legged robots are more complex to design, build, and
control~\citep{Raibert08,Roennau10,Semini11,johnson2015team} but they have obvious
mobility advantages when operating in unstructured terrains and environments.
Some research groups have started investigating mobile robot designs which
combine the advantages of both legged and wheeled locomotion, using different
coupling mechanisms between the wheels and legs~\citep{Adachi99,Endo00,Halme03}.

Recently, the DRC accelerated the
development of new mobile manipulation platforms aimed to address disaster
response tasks and search and rescue~(SAR) operations. While the majority of the
teams participating in the DRC Finals designed purely bipedal
robots\footnote{\url{http://web.archive.org/web/20160416073534/http://theroboticschallenge.org/teams} (original unavailable)}, four of the
five best placed teams chose to combine legged with wheeled locomotion, which
might indicate advantages of this design approach for the challenge tasks.
On the one hand, these
robots can move fast over flat terrain using their wheels, on the other hand,
they are able to overcome more complex terrain using stepping.

DRC-HUBO of the winning team KAIST is a humanoid
robot~\citep{cho2011,kim2010} capable of bipedal
walking. %
Its powerful joint motors are equipped with an air cooling system to dispense heat
efficiently and allow high payloads. DRC-HUBO can rotate its upper body
by $180^\circ$ which enables it to climb a ladder with the knees extending
backwards~\citep{lim2015}.
DRC-HUBO is also able to drive over flat terrain, using
wheels which are attached to its knees and ankles. To switch between
walking and driving, DRC-HUBO transforms between a standing posture
and a kneeling posture.

Team IHMC~\citep{johnson2015team} came in second at the DRC Finals and, from the five
best placed teams, was the only team using a purely bipedal robot with no additional
wheels or tracks: the Atlas robot developed by Boston Dynamics.

CHIMP~\citep{stentz2015chimp}, which placed 3rd in the DRC Finals, was designed
to maintain static stability---avoiding engineering challenges that arise if
complex balancing control techniques are needed to maintain dynamic stability.
The roughly anthropomorphic robot is equipped with powered tracks on
its arms and legs, which can be used to drive over uneven terrain. During
manipulation tasks, CHIMP rests on the two tracks of its hind legs, which still
provide mobility on flat terrain. Raising its frontal limbs allows the robot
to use its grippers to manipulate
objects. In contrast to our concept, CHIMP does not execute any stepping motions
to overcome bigger obstacles like stairs, but instead drives over them on its
four tracks while maintaining a low center of mass (COM). The user
interface of CHIMP combines manual and autonomous control, for example by
previewing candidate free-space motions to the operator.

Likewise, RoboSimian is a statically stable quadrupedal robot with an ape-like
morphology~\citep{satzinger2014experimental,hebert2015mobile}. It is equipped
with four generalized limbs, which can be used for locomotion and manipulation,
consisting of seven joints each. All of these 28 joints are driven by identical
actuators to ease development and maintenance of the robot hardware.
Furthermore, it is equipped with under-actuated hands at the end of its limbs
with fewer fingers and active DOF than a human hand. Besides executing stepping
motions with its limbs, it is also capable of driving on four wheels. For this
purpose, RoboSimian can lower itself onto two active wheels attached to its trunk
and two caster wheels on two of its limbs. This allows the robot to drive on
even terrain, while still being able to manipulate objects using its other two
limbs. RoboSimian placed 5th in the competition.

In contrast to DRC-HUBO, CHIMP, and RoboSimian, our robot Momaro is capable of
driving omnidirectionally, which simplifies navigation in restricted spaces
and allows us to make small positional corrections faster.
Furthermore, our robot is equipped with six limbs, two of which are exclusively
used for manipulation. The use of four legs for locomotion provides a large and
flexible support polygon when the robot is performing mobile manipulation tasks.

We developed a telemanipulation interface for our robot
using an immersive 3D HMD (Oculus Rift) and two 6D controllers (Razer Hydra), allowing an operator to intuitively
manipulate objects in the environment. Telemanipulation interfaces using 3D perception 
and a HMD have been addressed by multiple groups, for example for SAR robots~\citep{martins2009immersive},
explosive ordnance disposal~\citep{kron2004disposal}, or in surgery~\citep{ballantyne2003vinci,hagn2010dlr}.
In contrast, our telemanipulation solution consists of low-cost consumer-grade
equipment.

The idea of using consumer-grade equipment for robotic applications is not new.
\Citet{kot2014utilization} used the Oculus Rift as well in
their mobile manipulation setup using a four-wheeled robot with a 3\,DOF arm.
Similarly, \citet{smith2009wiimote} used the low-priced Wiimote game controller
with an additional IR camera to track the position and orientation of the
operator hand. They use a minimum jerk human motion
model to improve the precision of the tracking and achieved good results for
minimally instructed users in a simple manipulation task. In contrast to the
Wiimote, which can only measure linear accelerations, the Razer Hydra is able to
determine absolute positions using a magnetic field.
Compared to the previous work on telemanipulation, we describe a system that can be
intuitively teleoperated by a human operator---even under degraded network communication---and is highly mobile by using a combination of legged and wheeled locomotion.

\pagebreak
\section{Mobile Manipulation Robot Momaro}
\label{sec:momaro}

Our mobile manipulation robot Momaro (see \cref{fig:momaro}) was specifically
designed for the requirements of the DRC. Besides the overall goal to solve all DRC
tasks, we specified additional design constraints:
A \textit{bimanual design} offers both the ability to perform complex or strenuous
manipulation tasks
which might be impossible using only one hand, and also adds redundancy for one-handed tasks.
Bimanual manipulation is also a long-standing interest of our research group,
particularly in context of service robotics \citep{stuckler2014increasing}.
A \textit{large support polygon} minimizes the need for
balance control, which might be challenging, \eg for bipedal robots.
\textit{Legs} offer the ability to step over or climb on obstacles.
A \textit{lightweight robot} is less dangerous and also easier to handle than heavy robots
requiring special moving equipment. The capability of \textit{omnidirectional movement}
allows faster and more precise
correction movements in front of manipulation tasks, when compared to, \eg a robot that
needs to turn in order to move sideways.
Finally, since our hardware engineering capacities were limited, we 
wanted to use \textit{off-the-shelf components} as much as possible.

\subsection{Kinematic Design}

\begin{figure}
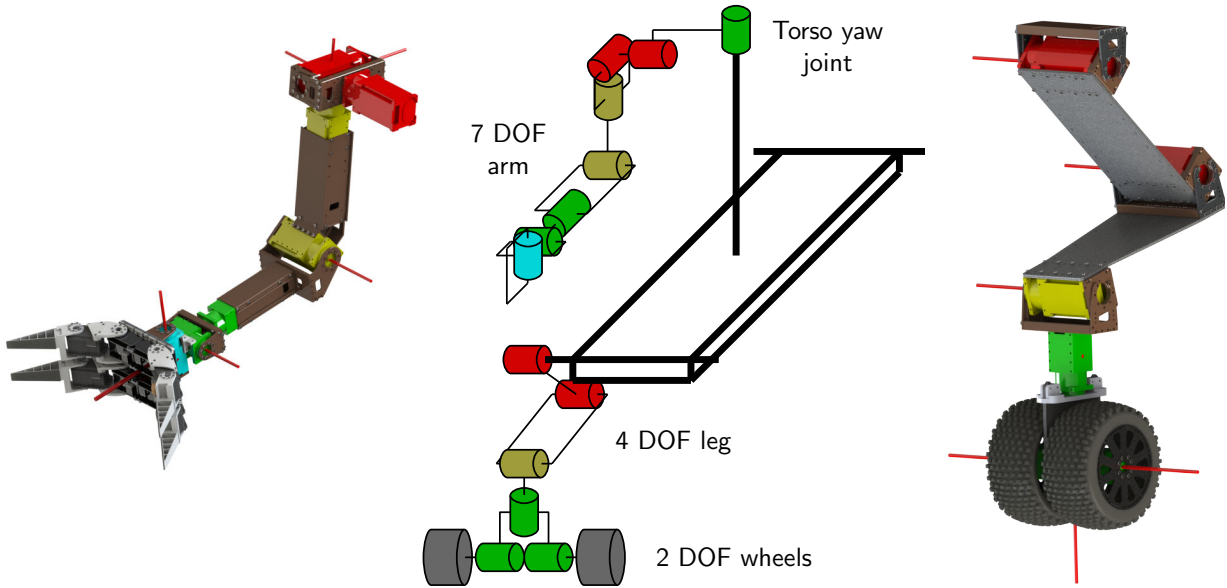

  \centering\begin{maybepreview}
  \includegraphics[trim=54mm 10mm 48mm 30mm,height=7.8cm,clip]{images/Arm_Hand_isometric_colored_kinematic_axes.png}%
\begin{tikzpicture}
    \colorlet{armcolor}{cyan!20}
    \colorlet{legcolor}{brown!20}
    \colorlet{wheelcolor}{yellow!20}

    \begin{pgfonlayer}{main}
    \node[anchor=south west,inner sep=0] (image) at (0,0) {\includegraphics[trim=6.5cm 2.6cm 7cm 1.5cm,height=7.8cm,clip]{images/Kinematic_chain_simple_servo_colors.pdf}};
    \end{pgfonlayer}
    \begin{scope}[
            x={(image.south east)},
            y={(image.north west)},
            font=\sffamily,
            every node/.style={align=center},
            line/.style={red, line width=1pt},
            box/.style={rectangle,draw=red,inner sep=0.3333em,line width=1pt}
          ]
    \node[anchor=west,rounded corners] at (0.45, 0.055) {2 DOF wheels};
    \node[anchor=west,rounded corners] at (0.37, 0.25) {4 DOF leg};
    \node[anchor=east,rounded corners] at (0.27, 0.75) {7 DOF\\arm};
    \node[anchor=west] at (0.68, 0.92) {Torso yaw\\joint};

    \begin{pgfonlayer}{background}
    \end{pgfonlayer}
    \end{scope}
\end{tikzpicture} %
  \includegraphics[trim=60mm 10mm 40mm 20mm,height=7.8cm,clip]{images/Coax_DynamixelPro_ass_flexi_forPaper.png}%
  \end{maybepreview}%
  \caption{%
    Kinematic layout.
    Left: CAD rendering of the right arm. Joint axes are marked with red lines.
    Center: Kinematic tree. For clarity, the figure only shows a part of the robot and does not show the hand with its additional eight DOF. Proportions are not to scale.
    Right: CAD rendering of the front right leg. The six joint axes in hip, knee, ankle pitch, ankle yaw, and wheels are marked with red lines.
  }
  \label{fig:Kinematic_chain}
\end{figure}

Driven by the off-the-shelf and lightweight design goals, we decided to
power all robot joints by Robotis Dynamixel actuators
(see~\cref{tab:actuators}), which offer a good torque-to-weight ratio.
Notably, all other high-placed DRC designs use custom actuator designs.
Figure~\ref{fig:Kinematic_chain} gives an overview of the kinematic structure of Momaro.

Since state of the art approaches for bipedal
locomotion on terrain are prone to falls and current generation robots are mostly not able
to recover after these falls by themselves, we decided to equip Momaro with a
total of four legs to minimize the probability of falling. As robot locomotion
using stepping is comparably slow, the legs end in pairs of steerable wheels.
The legs have three pitch joints
in hip, knee and ankle, allowing the adjustment of the wheel pair position
relative to the trunk in the sagittal plane. Furthermore, the ankle can rotate
around the yaw axis and the two wheels can be driven independently. This allows
the robot to drive omnidirectionally on suitable terrain, while also stepping
over obstacles too high to drive over.

The leg segments are carbon fiber springs, thus providing passive adaptation to terrain.
The foreleg extension varies 40\,cm from minimum to maximum, \ie from lowest to highest configuration of the robot.
In the minimum configuration, Momaro has a chassis clearance of 32\,cm.
The hind legs can extend 15\,cm more to allow the robot to climb steeper inclines while keeping the basis level.
While the legs can be used for locomotion,
they also extend the workspace of the robot for manipulation tasks, \eg by
changing the height of the base or by pitching/rolling the base through
antagonistic leg length changes.
The wheels are soft foam-filled rubber wheels, which provide ample traction.
Their radius of 8\,cm and the flexible suspension formed by
the carbon fiber springs allows the robot to ignore most obstacles lower
than approximately 5\,cm. %
Since our manipulation interfaces (see~\cref{sec:operator}) do not require
precise base positioning, the spring design does not decrease manipulation capabilities.
Additionally, unintended base movement is measured using the built-in IMU and compensated
for during sensor data processing (see~\cref{sec:perception}).

\begin{figure}
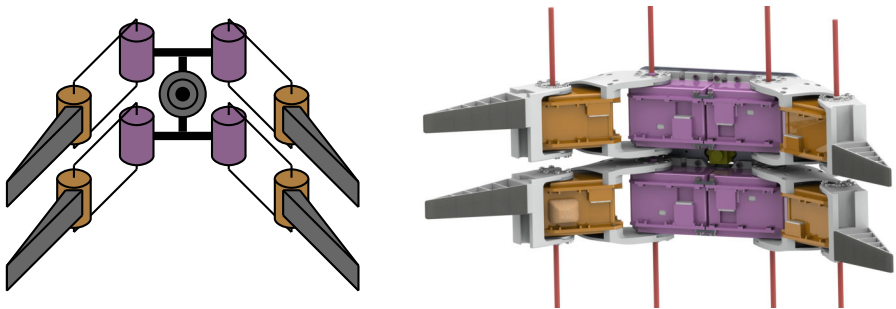

  \centering\begin{maybepreview}
  \includegraphics[trim=94mm 69mm 94mm 69mm,height=4cm,clip]{images/Kinematic_chain_hand_servo_colors.pdf}
  \includegraphics[trim=0 120 0 100,clip,height=4cm]{images/Hand_isometric_Camera_v05_with_Axes_colored.png}\end{maybepreview}
  \caption{%
    Gripper design.
    Left: Kinematic tree of one of Momaro's hands. While all segments connecting the joints are rigid, the distal finger segments deform if force is applied to them. Proportions are not to scale.
    The color camera mounted in the hand is visible in the center.
    Right: CAD rendering of the hand. The finger joint axes are marked with red lines.
  }
  \label{fig:robot:arm}
\end{figure}

On top of its flexible base, Momaro has an anthropomorphic upper body consisting
of two adult-sized, 7\,DOF arms (see \cref{fig:momaro,fig:Kinematic_chain}) and a sensor head.
The upper body of the robot
is connected to the base by a torso yaw joint that increases the workspace of
the end-effectors and allows the system to execute more tasks without the use of
locomotion. Each arm ends in a custom hand equipped with four 2 DOF fingers (see \cref{fig:robot:arm}).
While the proximal segment of each finger is rigid, Festo
FinGrippers are used as distal segments. These grippers deform if force is
applied to them to better enclose a grasped object by enlarging the contact
surface between object and gripper. The position of the finger tips on each
finger can manually be reconfigured to allow pinch grips as well as cylindrical
grasps.

Momaro is relatively lightweight (58\,kg) and compact (base footprint
80\,cm$\times$70\,cm), which means that it
can be carried comfortably by two people, compared to larger crews and
equipment like gantries needed to carry other robots of comparable size.
Since the legs and upper body can be detached,
the robot can be transported in standard suitcases.

\FloatBarrier

\begin{table}
  \centering
  \begin{threeparttable}
    \caption{%
      Robotis Dynamixel actuator models used in Momaro.
    }
    \label{tab:actuators}
    \begingroup
    \small
    \setlength{\tabcolsep}{5.1pt}
    \begin{tabular}{llrr|llrr}
      \toprule
      Joint         & Model            & Mass & Torque                                              & Joint         & Model              & Mass & Torque \\
      \midrule
      Hip           & \cellcolor{red!25}{H54-200-S500-R   } & 855\,g & 44.2\,Nm       & Shoulder (r.+p.) & \cellcolor{red!25}{2$\times$ H54-200-S500-R}  & 855\,g & 44.2\,Nm \\
      Knee          & \cellcolor{red!25}{H54-200-S500-R   } & 855\,g & 44.2\,Nm       & Shoulder (yaw)& \cellcolor{yellow!25}{H54-100-S500-R}     & 732\,g & 24.8\,Nm \\
      Ankle (pitch) & \cellcolor{yellow!25}{H54-100-S500-R   } & 732\,g & 24.8\,Nm    & Elbow         & \cellcolor{yellow!25}{H54-100-S500-R   }  & 732\,g & 24.8\,Nm \\
      Ankle (yaw)   & \cellcolor{green!25}{H42-20-S300-R    } & 340\,g & 6.3\,Nm      & Wrist (roll)  & \cellcolor{green!25}{H42-20-S500-R   }  & 340\,g & 6.3\,Nm \\
      Wheels        & \cellcolor{green!25}{2$\times$ H42-20-S300-R } & 340\,g & 6.3\,Nm      & Wrist (pitch) & \cellcolor{green!25}{H42-20-S300-R    }  & 340\,g & 6.3\,Nm \\
                    &                                         &        &              & Wrist (yaw)   & \cellcolor{cyan!25}{L42-10-S300-R    }  & 257\,g & 1.4\,Nm \\
      Torso (yaw)   & \cellcolor{green!25}{H42-20-S300-R    } & 340\,g & 6.3\,Nm      & Proximal fingers  & \cellcolor{violet!30}{4$\times$ MX-106        }  & 153\,g & 8.4\,Nm \\
      Laser         & \cellcolor{orange!50}{MX-64            } & 126\,g & 6.0\,Nm     & Distal fingers  & \cellcolor{orange!50}{4$\times$ MX-64         }  & 126\,g & 6.0\,Nm \\
      \bottomrule
    \end{tabular}
    \endgroup
    \footnotesize The colors match the actuator colors in \cref{fig:Kinematic_chain,fig:robot:arm}.
  \end{threeparttable}
\end{table}

\subsection{Sensing}

\begin{figure}
  \centering\begin{maybepreview}
\begin{tikzpicture}
    \node[anchor=south west,inner sep=0] (image) at (0,0) {\includegraphics[height=4.5cm]{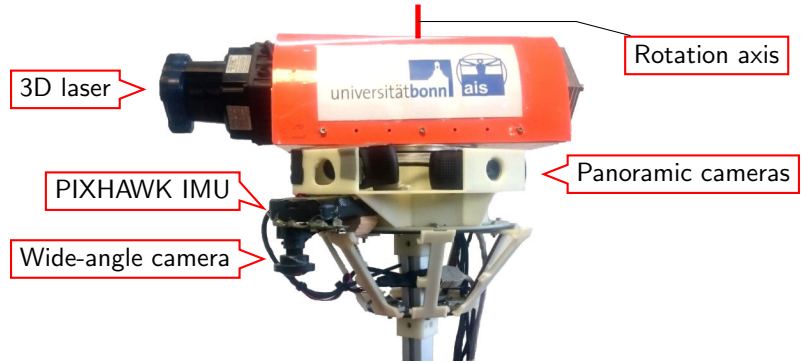}};
    \begin{scope}[
            x={(image.south east)},
            y={(image.north west)},
            font=\sffamily,
            every node/.style={align=center,text height=1.5ex,
    text depth=.25ex,
    text centered
    },
            line/.style={red, line width=1pt},
            box/.style={rectangle,draw=red,inner sep=0.3333em,thick}
          ]
    \node[box,anchor=east,rectangle callout, callout relative pointer={(0.05,0.0)}] at (-0.05, 0.8) {3D laser};
    \node[box,anchor=west,rectangle callout, callout relative pointer={(-0.05,0.0)}] at (0.88, 0.55) {Panoramic cameras};
    \node[box,anchor=east,rectangle callout, callout relative pointer={(0.05, 0.0)}] at (0.2, 0.3) {Wide-angle camera};
    \node[box,anchor=east,rectangle callout, callout relative pointer={(0.05, -0.01)}] at (0.2, 0.5) {PIXHAWK IMU};    

    \draw[line width=2pt,red] (0.57, 1.05) -- (0.57, 0.95);

    \node[box,anchor=west] (axis) at (1.0, 0.9) {Rotation axis};
    \draw (axis) -- (0.9,1.0) -- (0.57,1.0);
    \end{scope}
\end{tikzpicture} %
\end{maybepreview}
  \caption{Sensor head carrying 3D laser scanner, IMU, and panoramic cameras.}
  \label{fig:sensor_head}
\end{figure}

Momaro's main sensor for environmental perception is a 3D rotating laser scanner
on its sensor head (see~\cref{fig:sensor_head}).
It consists of a Robotis Dynamixel MX-64 actuator, which rotates a Hokuyo
UTM-30LX-EW laser scanner around the vertical axis. A PIXHAWK IMU is mounted close
to the laser scanner, which is used for motion compensation during scan
aggregation and state estimation.
Three Full HD color cameras are also attached to the sensor
head for a panoramic view of the environment in front of the robot and a
top-down wide angle camera is used to observe the movement of the arms of the
robot and its interaction with the environment. Each hand is equipped with a
camera which is located between its fingers. These cameras can be used to
visually verify the correct grasp of objects. Furthermore, since these cameras
are mounted at the end-effectors of the robot and can therefore be moved, they
can be used to extend the view of the operators, for example, to view a scene
from another perspective if the view from the head mounted top-down camera is
occluded.
Finally, the robot also carries a downward-facing wide-angle camera under its
base which allows the operators to monitor the wheels and the surface beneath Momaro.

Since the right hand is used for the more complex tasks, it is equipped with
additional sensors. A microphone connected to the hand camera can be used
for auditory feedback to the operators. Underneath the hand, we mounted an
infrared distance sensor to measure distances within the environment.

\subsection{Electronics}

\begin{figure}
  \centering\begin{maybepreview}
  \includegraphics[width=\linewidth]{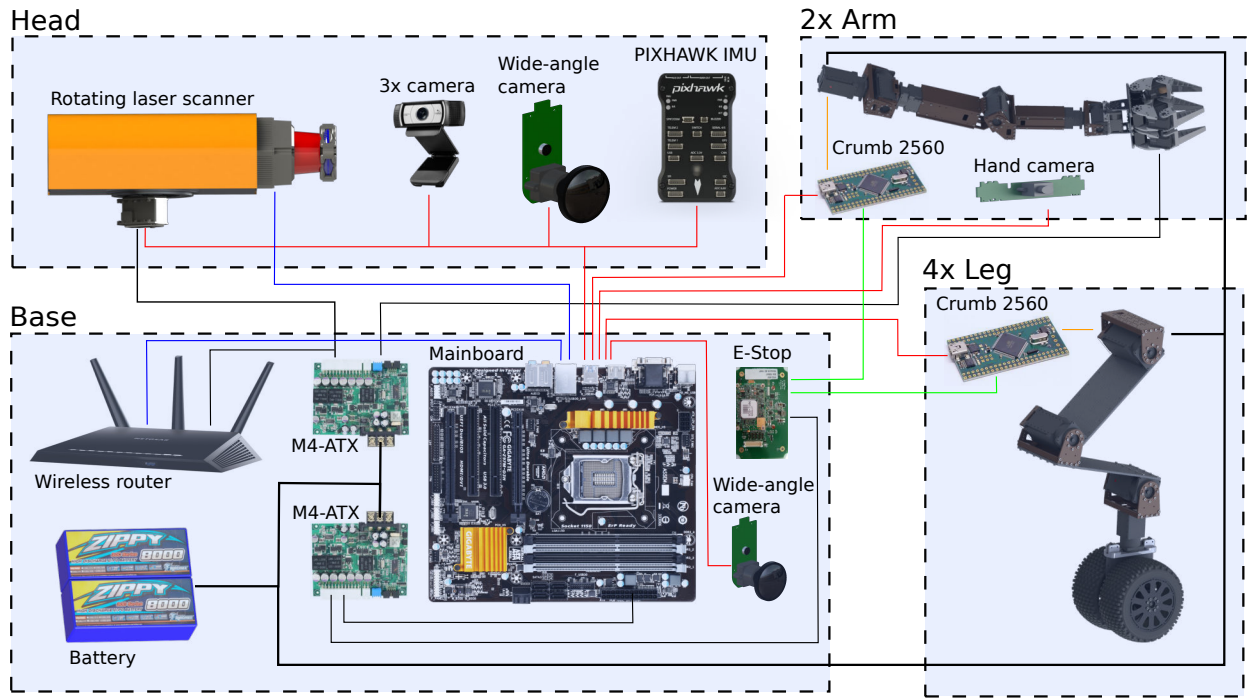}\end{maybepreview}
  \caption{%
    Simplified electrical schematics of Momaro.
    We show USB 2.0 data connections (red), LAN connections (blue), E-Stop related wiring (green), the low-level servo bus system (orange), and power connections (black).
    Thick black lines indicate battery power, whilst thin black lines carry 12\,V.}
  \label{fig:Momaro_electrical_sceme}
\end{figure}

Figure~\ref{fig:Momaro_electrical_sceme} shows an overview over the electrical components of Momaro. In its base, Momaro carries an on-board computer with a fast CPU (Intel Core i7-4790K @4--4.4\,GHz) and 32\,GB RAM. For communication with the operator station, it is equipped with a NETGEAR Nighthawk AC1900 WiFi router, which allows 2.4\,GHz and 5\,GHz transmission with up to 1300\,Mbit/s.
We make use of a total of six (one for each leg and arm) Crumb2560 microcontroller boards, which bridge high-level USB commands from the computer to low-level servo control commands and vice versa. Performance of the joint actuators is continuously monitored. Feedback
information includes measured position, applied torque, and actuator temperature.
Like the microcontroller boards, all cameras, the servo for rotation of the laser, and the PIXHAWK IMU are connected via USB 2.0 for a total of 16 USB devices. The laser scanner is connected via 100\,Mbit/s LAN through a slip ring.

In case of undesirable actions or emergencies, Momaro can be emergency-stopped through two emergency stop switches. One is mounted on the base of the robot for easy access during development, the other one is the wireless E-Stop system mandatory for all DRC competitors. The E-stops are connected to the actuator control microcontrollers. If the robot is E-stopped, it stops all currently active servo commands.

Power is supplied to the robot by a six-cell LiPo battery with 16\,Ah capacity at 22.2\,V nominal voltage, which yields around 1.5--2\,h run time, depending on the performed tasks. Batteries are hot-swappable and thus can be easily exchanged while running. For comfortable development and debugging, they can also be substituted by a power supply.

\subsection{Design Iterations}

Shortly before the DRC Finals, DARPA announced a reduced set of tasks.
One manipulation task (cylindrical plug) was removed and two terrain tasks
were combined into one. This increased the importance of the car tasks in relation to the
manipulation tasks. While initially our plan was to neglect the car task in favor of
the manipulation tasks, we now knew that we had to prepare for it, despite having
no car to practice with.
We decided to reduce the height of the base box
and to increase the length of the hind legs. Both measures allowed the robot to touch the
ground with the hind legs when sitting in the car. In the end, the modifications were
finished with few days left before the competition.
This base iteration was the only major change we made---apart from it the robot
performed in the DRC as initially designed.
\section{Perception}
\label{sec:perception}

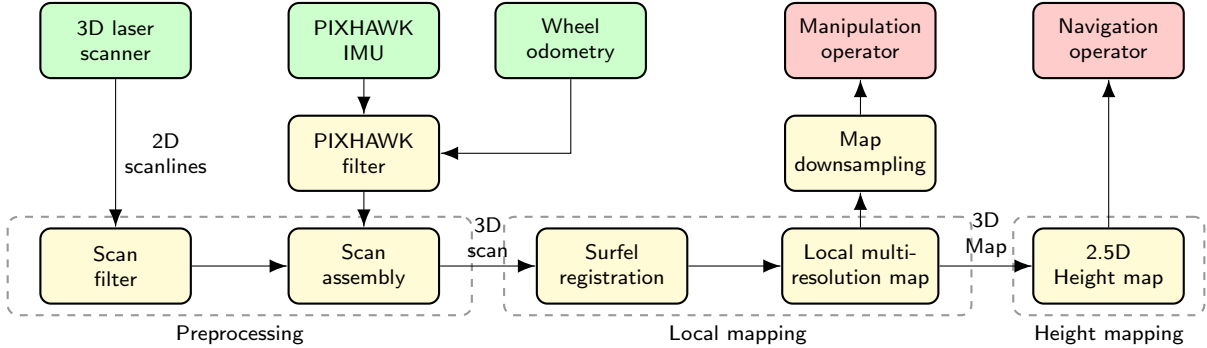
\begin{figure}
\centering\begin{maybepreview}
\begin{tikzpicture}
    [
    font=\sffamily\footnotesize,
    n/.style={draw, thick, rounded corners,fill=yellow!20,minimum height=1cm, minimum width=2cm},
    sensor/.style={n,fill=green!20},
    op/.style={n,fill=red!20},
    module/.style={dashed,draw=black!40,thick,rounded corners,inner sep=0.3cm,transform shape,inner sep=4pt},
    every node/.style={
      align=center
    },
    xscale=1.1
  ]

  \node[n] (scanfilter) at (0.0, 0.0) {Scan\\filter};
  \node[n] (assembler) at (3.0, 0.0) {Scan\\assembly};
  \node[n] (registration) at (6.0, 0.0) {Surfel\\registration};
  \node[n] (map) at (9.0, 0.0) {Local multi-\\resolution map};
  \node[n] (heightmap) at (12.0, 0.0) {2.5D\\Height map};

  \node[module,fit={(scanfilter) (assembler)}] (preprocessing) {};
  \node[below = 0.0cm of preprocessing] {Preprocessing};

  \node[module,fit={(registration) (map)}] (local_mapping) {};
  \node[below = 0.0cm of local_mapping] {Local mapping};

  \node[module,fit={(heightmap)}] (height_mapping) {};
  \node[below = 0.0cm of height_mapping] {Height mapping};

  \draw[big arrow] (scanfilter) -- (assembler);
  \draw[big arrow] (assembler) -- (registration) node [midway, above] {3D\\scan};
  \draw[big arrow] (registration) -- (map);%
  \draw[big arrow] (map) -- (heightmap) node[midway,above] {3D\\Map};

  \node[sensor] (laser) at (0.0, 3.0) {3D laser\\scanner};
  \draw[big arrow] (laser) -- (scanfilter) node [midway, right,font=\footnotesize\sffamily] {2D\\scanlines};

  \node[sensor] (imu) at (3.0, 3.0) {PIXHAWK\\IMU};
  \node[sensor] (odom) at (5.5, 3.0) {Wheel\\odometry};
  \node[n] (imufilter) at (3.0, 1.5) {PIXHAWK\\filter};

  \draw[big arrow] (imu) -- (imufilter);
  \draw[big arrow] (odom) |- (imufilter);
  \draw[big arrow] (imufilter) -- (assembler);

  \node[n] (downsampling) at (9.0, 1.5) {Map\\downsampling};
  \draw[big arrow] (map) -- (downsampling);

  \node[op] (manip_op) at (9.0, 3.0) {Manipulation\\operator};
  \draw[big arrow] (downsampling) -- (manip_op);

  \node[op] (nav_op) at (12.0, 3.0) {Navigation\\operator};
  \draw[big arrow] (heightmap) -- (nav_op);
\end{tikzpicture}
\end{maybepreview}
\caption{Overview of our 3D laser perception system. The measurements are processed in preprocessing steps described in \cref{sec:scan_assembly}. The resulting 3D point cloud is used to estimate the transformation between the current scan and the map. Registered scans are stored in a local multiresolution map.}
\label{fig:perception_architecture}
\end{figure}

To assist the operators in navigation and manipulation tasks, we construct a 3D egocentric local multiresolution grid map by accumulating laser range measurements that
are made in all spherical directions. The architecture of our perception and mapping system is outlined in \cref{fig:perception_architecture}.
3D scans are acquired in each full rotation of the laser.
Since a rotation takes time, the motion of the robot needs to be compensated when assembling the scan measurements into 3D scans (\cref{sec:scan_assembly}).
We first register newly acquired 3D scans with the so far accumulated map and then update the map with the registered 3D scan to estimate the motion of the robot, compensating for drift of the wheel odometry and IMU measurements.

\subsection{Preprocessing and 3D Scan Assembly}\label{sec:scan_assembly}

The raw measurements from the laser scanner are subject to spurious measurements at occluded transitions between two objects. These so-called \textit{jump edges} are filtered by comparing the angle of neighboring measurements. After filtering for jump edges, we assemble a 3D scan from the 2D scans of a complete rotation of the scanner. Since the sensor is moving during acquisition, we undistort the individual 2D scans in two steps.

First, measurements of individual 2D scans are undistorted with regards to the rotation of the 2D laser scanner around the sensor rotation axis. Using spherical linear interpolation, the rotation between the acquisition of two scan lines is distributed over the measurements.

Second, the motion of the robot during acquisition of a full 3D scan is compensated. Due to Momaro's flexible legs, it is not sufficient to simply use wheel odometry to compensate for the robot motion. Instead, we estimate the full 6D state with the PIXHAWK IMU attached to Momaro's head. Here we calculate a 3D attitude estimate from accelerometers and gyroscopes to compensate for rotational motions of the robot. Afterwards, we filter the wheel odometry with measured linear acceleration to compensate for linear motions. The resulting 6D state estimate includes otherwise unobservable motions due to external forces like rough terrain, contacts with the environment, wind, etc. It is used to assemble the individual 2D scans of each rotation to a 3D scan.

\subsection{Local Multiresolution Map}\label{sec:multimap}

\begin{figure}
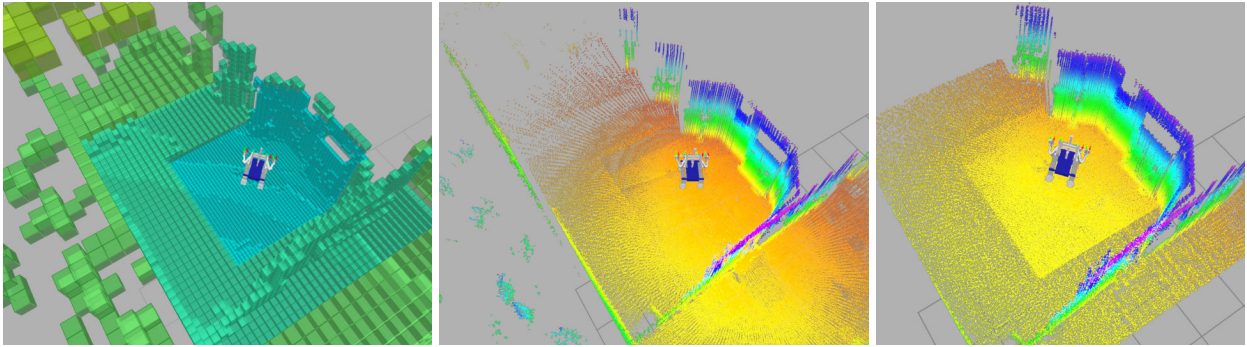

\centering\begin{maybepreview}
\includegraphics[height=0.275\linewidth]{images/perception/gridmap.png}
\includegraphics[height=0.275\linewidth]{images/perception/pointcloud.png}
\includegraphics[height=0.275\linewidth]{images/perception/pointcloud_downsampled.png}\end{maybepreview}
\caption{The local multiresolution grid-based map during the first DRC competition run. Left: The grid-based local multiresolution map.
Cell size (indicated by color) increases with the distance from the robot. Middle: 3D points stored in the map on the robot.
Right: Downsampled and clipped local map, transmitted to the operator for manipulation and navigation tasks. Color encodes height above ground. }  
\label{fig:perception_map}
\end{figure}

The assembled 3D scans are accumulated in a hybrid local multiresolution grid-based map. Measurements and occupancy information are stored in grid cells that increase in size with the distance from the robot center. The individual measurements are stored in ring buffers enabling constant size in memory. More recent measurements replace older measurements. 
By using multiresolution, we gain a high measurement density in the close proximity to the sensor and a lower measurement density far away from our robot, which correlates with the sensor characteristics in relative distance accuracy and measurement density.
Compared to uniform grid-based maps, multiresolution leads to the use of fewer grid cells, without losing relevant information and consequently results in lower computational costs. \cref{fig:perception_map} shows an example of our grid-based map.

Maintaining the egocentric property of the map necessitates efficient map management for translation and rotation during motion. Therefore, individual grid cells are stored in ring buffers to allow shifting of elements in constant time. Multiple ring buffers are interlaced to obtain a map with three dimensions. In case of a translation of the robot, the ring buffers are shifted whenever necessary. For sub-cell-length translations, the translational parts are accumulated and shifted if they exceed the length of a cell. 

Newly acquired 3D scans are aligned to the local multiresolution map by our surfel registration method~\citep{droeschel2014local}. We gain efficiency by summarizing individual points in each grid cell by a sample mean and covariance.

\subsection{Height Mapping}\label{sec:heightmapping}

Besides assisting the operators for navigation and manipulation tasks, the local map is used by the autonomous stepping module to plan footsteps. To this end, the 3D map is projected into a 2.5D height map, shown in~\cref{fig:stepping}.
Gaps in the height map (cells without measurements) are filled with the local
minimum if they are within a distance threshold of valid measurements
(10\,cm in our experiments). The rationale for using the local minimum is that
gaps in the height map are usually caused by occlusions. The high mounting
position of the laser on the robot means that low terrain is more likely
occluded than high terrain. The local minimum is therefore a
good guess of missing terrain height.
After filling gaps in the height map, the height values are filtered using the
fast median filter approximation using local histograms~\citep{huang1979}.
The filtered height map is suitable for planning footsteps. %

\section{Communication}

One constraint during the DRC was the limited communication between the operator
station and the robot, which was enforced to simulate degenerated communication
as may occur in a real-world mission. The uplink from the operator station to
the robot was limited to 9600\,bit/s at all times. The downlink from the robot
to the operator station was limited to 300\,Mbit/s outside of the building
during the driving tasks, the door task, and the stairs task. Inside the
building (defined by the door
thresholds), the downlink was limited to 9600\,bit/s, interleaved with one
second long bursts of 300\,Mbit/s bandwidth. These burst became more
frequent during the run and the blackouts vanished completely after 45 minutes
into the run. As usual, the
wireless communication link does not guarantee packet delivery, so communication
systems had to deal with packet loss.

To cope with this degraded communication, sensor information cannot be
transferred unselected and uncompressed. The main idea of our communication
system is to transfer stationary information about the environment over the
high-latency high bandwidth channel, while we use the low-latency low bandwidth
channel to transfer frequently changing data. Both are then combined on the
operator station to render immersive 3D visualizations with low latency for the
operators.

\subsection{Communication Architecture}

\begin{figure}[tbp]
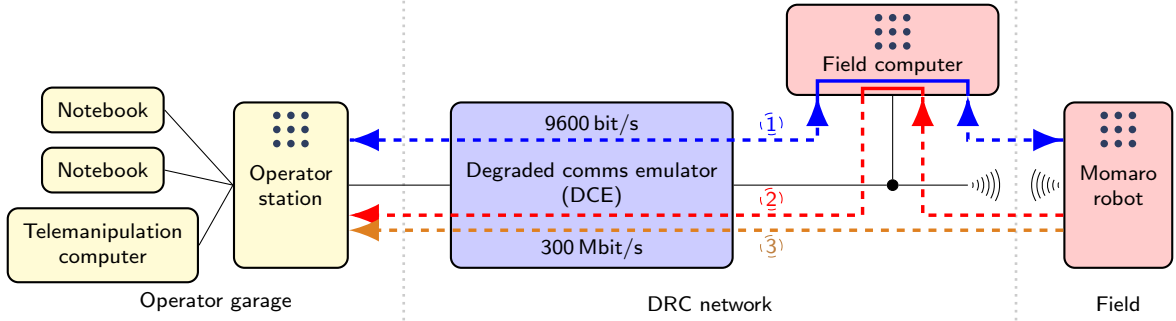

  \centering\begin{maybepreview}
\begin{tikzpicture}
    [
    font=\sffamily\footnotesize,
    pc/.style={draw,thick,rounded corners,fill=yellow!20,inner sep=.2cm,align=center},
    inf/.style={draw,thick,rounded corners,fill=blue!20,inner sep=.2cm,align=center},
    robot/.style={draw,thick,rounded corners,fill=red!20,inner sep=.2cm,align=center},
    to/.style={>=stealth',shorten >=1pt,semithick,font=\sffamily\footnotesize},
    transmission/.style={decorate, decoration={expanding waves, angle=30,
                          segment length=2}},
    stream/.style={dashed,line width=1.3pt},
    boundary/.style={dotted,line width=1pt,draw=black!20},
  ]
  \pgfmathsetmacro{\labely}{-1.8}

    \coordinate (b_op_drc) at (-2.5, \labely);
    \coordinate (b_drc_rob) at (5.64, \labely);

    \draw[boundary] (b_op_drc) -- ($(-10, 2.5)!(b_op_drc)!(10,2.5)$);
    \draw[boundary] (5.64, -1.8) -- (5.64,2.5);
    \node[anchor=south] at (-5, \labely) {Operator garage};
    \node[anchor=south] at ($(b_op_drc)!0.5!(b_drc_rob)$) {DRC network};
    \node[anchor=south] at (7, \labely) {Field};

    \node[pc] (pc1) at (-6.5, -0.8) {Telemanipulation\\computer};
    \node[pc] (pc2) at (-6.5,  0.2) {Notebook};
    \node[pc] (pc3) at (-6.5,  1) {Notebook};

    \node[pc,minimum height=2.2cm] (operator) at (-4, 0) {Operator\\station};

    \node[inf,minimum height=2.2cm] (ne) at (0, 0) {Degraded comms emulator\\(DCE)};

    \node[circle,draw=black,fill=black,inner sep=0.05cm] (switch) at (4, 0) {};
    \coordinate (wifi) at (5, 0);

    \node[robot,minimum height=2.2cm] (robot) at (7, 0) {Momaro\\robot};

    \node[robot,minimum height=1.2cm,minimum width=8em] (field) at (4, 1.8) {};
    \node[anchor=south] at ($(field.south) + (0,0.15)$) {Field computer};

    \draw (pc1.east) -- (operator.west);
    \draw (pc2.east) -- (operator.west);
    \draw (pc3.east) -- (operator.west);

    \draw (operator) -- (ne);

    \draw (ne) -- (switch);

    \draw (switch) -- (field);
    \draw (switch) -- (wifi);

    \draw[transmission] (robot) -- ($(wifi.east) + (0.8,0)$);
    \draw[transmission] (wifi) -- ($(robot.west) + (-0.8,0)$);

    \draw[stream,red,big arrow] ($(robot.west) + (0,-0.4)$) -| ($(field.south) + (0.4,0.0)$);
    \draw[stream,solid,red] ($(field.south) + (0.4,0.0)$) |- ($(field.south) + (0.0, 0.1)$) -| ($(field.south) + (-0.4, 0.0)$);
    \draw[stream,red,big arrow] ($(field.south) + (-0.4,0.0)$) |- ($(operator.east) + (0.0,-0.4)$);
    \node[anchor=south,red,circle,draw=red,inner sep=0.5,outer sep=2.0,dashed] at ($(ne.east) + (0.5,-0.4)$) {2};
    \node[anchor=north,inner sep=0.1cm] at (0,-0.6) {300\,Mbit/s};

    \draw[stream,blue,big 2arrow] ($(robot.west) + (0.0,0.6)$) -| ($(field.south) + (1.0, 0.0)$);
    \draw[stream,solid,blue] ($(field.south) + (1.0,0.0)$) |- ($(field.south) + (0.0, 0.2)$) -| ($(field.south) + (-1.0, 0.0)$);
    \draw[stream,blue,big 2arrow] ($(field.south) - (1.0, 0.0)$) |- ($(operator.east) + (0.0,0.6)$);
    \node[anchor=south,blue,circle,draw=blue,inner sep=0.5,outer sep=2.0,dashed] at ($(ne.east) + (0.5,0.6)$) {1};
    \node[anchor=south,inner sep=0.05cm] at (0,0.6) {9600\,bit/s};

    \draw[stream,brown!50!orange,big arrow] ($(robot.west) + (0.0, -0.6)$) -- ($(operator.east) + (0.0,-0.6)$);
    \node[anchor=north,brown,circle,draw=brown,inner sep=0.5,outer sep=2.0,dashed] at ($(ne.east) + (0.5,-0.6)$) {3};

  \node[anchor=north] at (operator.north) {\includegraphics[width=0.5cm]{images/ros_logo.pdf}};
  \node[anchor=north] at (field.north) {\includegraphics[width=0.5cm]{images/ros_logo.pdf}};
  \node[anchor=north] at (robot.north) {\includegraphics[width=0.5cm]{images/ros_logo.pdf}};
\end{tikzpicture} %
\end{maybepreview}
  \caption{%
    Communication architecture.
    Components in the vicinity of the operators are shown in yellow,
    DARPA-provided components in blue, components in the ``field''-network in red.
    Solid black lines represent physical network connections. Dashed lines show
    the different channels, which stream data over the network
    (blue\,(1): low bandwidth, red\,(2): bursts, brown\,(3): direct imagery).
    The ROS logo ($\vcenter{\hbox{\protect\includegraphics[height=\baselineskip]{images/ros_logo.pdf}}}$) indicates a ROS master.}
  \label{fig:commlink}
\end{figure}

Our communication architecture is shown in \cref{fig:commlink}. The main
topology was formed by the DARPA requirements, which placed the Degraded
Communications Emulator (DCE) between the operator crew and the robotic system.
To allow buffering and relaying of data over the high-bandwidth link, we
make use of the option to include a separate field computer, which is connected
via Ethernet to the DCE on the robot side. The key motivation here is that
the wireless link to the robot is unreliable, but unlimited in bandwidth,
while the link over the DCE is reliable, but limited in bandwidth. Placing the
field computer directly after the DCE allows exploitation of the characteristics
of both links.

On the operator side of the DCE, the operator station is the central unit.
Since our operator crew consists of more than one person, we have the option
to connect multiple specialist's notebooks to the operator station. Finally,
the computer running our telemanipulation interface (see \cref{sec:operator:razer}) is also
directly connected to the operator station.
Since we use the ROS middleware for all software components, separate
ROS masters run on the robot, the field computer, and the operator station.
The communication between these masters can be split into three channels, which
will be explained below.

\subsubsection{Low-bandwidth Channel}

The low-bandwidth channel is a bidirectional channel between the operator
station and robot (blue\,(1) in \cref{fig:commlink}).
It uses the low-bandwidth link of the DCE and is therefore always available.
Since the bandwidth is very limited, we do most compression
on the field computer, where we can be sure that packets sent to the operator station
are not dropped, which would waste bandwidth.

\begin{table}
\centering
\begin{threeparttable}
  \caption{Average bit rates of topics transmitted over the low-bandwidth link.}
  \label{tab:lowband_rates}
  \begin{tabular}{lrrlrr}
    \toprule
    \multicolumn{3}{c}{Robot $\rightarrow$ Operator}  & \multicolumn{3}{c}{Operator $\rightarrow$ Robot}   \\
    \cmidrule(lr){1-3} \cmidrule(lr){4-6}
    Channel/Topic      & Rate      & Avg. Bit/message & Channel/Topic       & Rate  & Avg. Bit/message     \\
    \midrule
    H.264 Camera image & 1\,Hz     & 6000             & Arm control         & 5\,Hz & 96                   \\
    Joint positions    & 1\,Hz     & 736              & Joystick command    & 5\,Hz & 56                   \\
    Base status        & 1\,Hz     & 472              & Generic motion\tnote{1} & - & 144                  \\
    3D Contour points     & 1\,Hz     & 250              & Motion play request & - & 80                   \\
    Transforms         & 1\,Hz     & 136              &                     &       &                      \\
    Audio amplitude    & 1\,Hz     & 8                &                     &                              \\
    \midrule
    \textbf{Sum per 1s}&           & 7602             &                     &       & -\tnote{2} \\
    \bottomrule
  \end{tabular}
  \vspace{0.5em}
  \footnotesize Topics with rate of ``-'' are transmitted only on operator request.
  \begin{tablenotes}
    \item [1] Generic transport for all kinds of keyframe motions. Here: one frame using Cartesian EEF pose.
    \item [2] Summation is not applicable here, since the total bit rate depends heavily on operator action.
  \end{tablenotes}
\end{threeparttable}
\end{table}

Since the low-bandwidth link over the DCE was the main live telemetry source for
the operator crew, we spent considerable effort on compressing the data sent
over this link in order to maximize the amount of information the system
provides. The transmitter running on the field computer sends exactly one
UDP packet per second. The bandwidth is thus easily controlled by limiting
the UDP payload size. Since the amount of data is much less in the other
direction, the transmitter on the operator station sends operator commands
with up to 5\,Hz.
Payload sizes in bits are given in \cref{tab:lowband_rates}.

For low-level compression of floating point numbers as well as 3D/4D vectors
and quaternions, we developed a small helper library, which is freely
available\footnote{\url{https://github.com/AIS-Bonn/vector_compression}}.
It employs techniques originally developed for compressing geometry data for
transfers between CPU and GPU: Quaternions are compressed using the hemi-oct
encoding \citep{cigolle2014survey}, while 3D vectors are compressed using
a face-centered cubic packing lattice. The lattice approach offers better
average discretization error than naive methods which discretize each axis
independently.

\begin{figure}
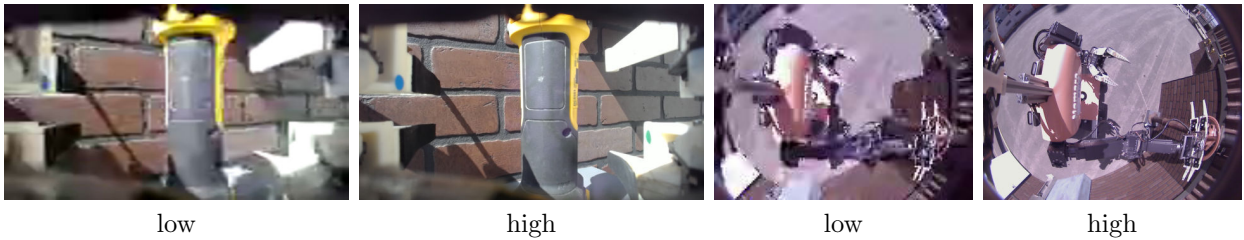

 \centering\begin{maybepreview}
 \begingroup
 \setlength{\tabcolsep}{2pt}
 \begin{tabular}{cccc}
 \includegraphics[height=2.6cm]{images/lowbandwidth/tool2/low_corrected.png} &
 \includegraphics[height=2.6cm]{images/lowbandwidth/tool2/high.png} &
 \includegraphics[height=2.6cm]{images/lowbandwidth/overhead3/low.png} &
 \includegraphics[height=2.6cm]{images/lowbandwidth/overhead3/high.png} \\
 low & high & low & high
 \end{tabular}
 \endgroup\end{maybepreview}
 \caption{Comparison of webcam images over low- and high-bandwidth channels.
  The left two images were captured by the right hand
  camera, looking at the drill tool. The right two images show the overhead view
  while the robot is grasping the valve.}
 \label{fig:camera_comparison}
\end{figure}

Since visual information is of crucial importance to human operators, we also
transmit a low resolution video stream. As Momaro is equipped with a variety of
cameras, an operator needs to select the camera whose output should be sent over
the low bandwidth link. The selection of the camera depends on the currently
executed task and is also often changed during a task.
Note that all camera images are also transmitted over the high-bandwidth link.
The purpose of low-bandwidth imagery is merely to provide low-latency feedback
to the operators.
The selected camera image is compressed at the field PC
using the H.264 codec. Before compression, the image is downscaled to
160$\times$120 pixels. Furthermore, we use the periodic intra refresh technique instead of
dedicated keyframes, which allows to specify a hard packet size limit for each
frame. While the compression definitely reduces details (see
\cref{fig:camera_comparison}), the camera
images still allow the operators to make fast decisions without waiting
for the next high-bandwidth burst image.

Measured joint positions are discretized and transmitted as 16\,bit integers
(8\,bit for the most distal joints in the kinematic tree).
The joint positions are used for
forward kinematics on the operator station to reconstruct poses of all robot
parts. A small number of 3D rigid body transformations are sent over the
network, including the current localization pose, odometry, and IMU information.
The transforms are sent as 3D vector and quaternion pairs, compressed using the
library mentioned above.

\begin{figure}
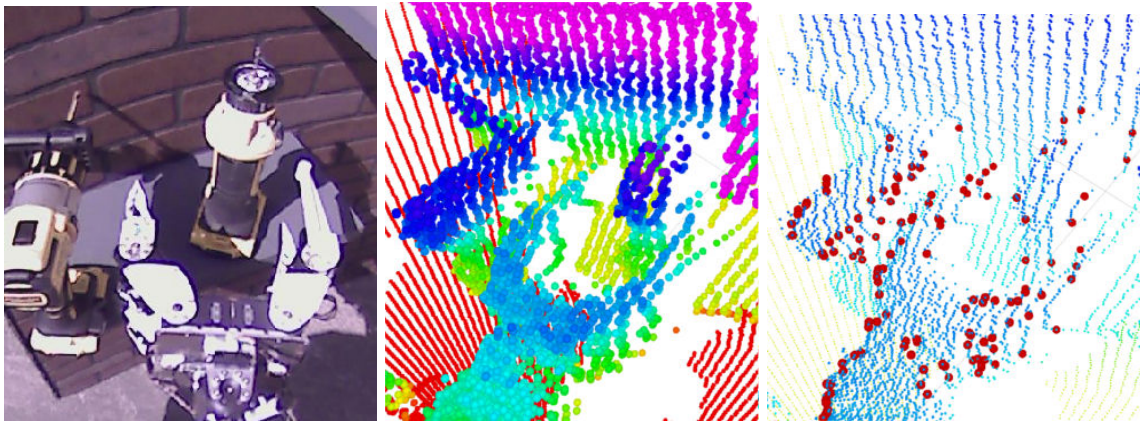

 \centering\begin{maybepreview}
 \includegraphics[width=0.3\linewidth]{images/interest_points_overhead.png}
 \includegraphics[width=0.3\linewidth]{images/interest_points_pointcloud.png}
 \includegraphics[width=0.3\linewidth]{images/interest_points_output.png}\end{maybepreview}

 \caption{3D contour points for a typical manipulation task (grasping the drill).
  Left: The overhead camera image. Middle: The raw laser scans (color encodes height from ground). Right: The resulting contour points (red). }
 \label{fig:interest_points}
\end{figure}

Up to 125 3D \textit{contour points} are compressed and sent to the operator for display.
These contour points are extracted from the laser scans and are meant to outline the contour of the endeffector and objects in its direct vicinity.
By transmitting contour points over the low-bandwidth channel, the operator is provided with live sensory feedback from the laser scanner during a manipulation task.
\Cref{fig:interest_points} shows the extracted contour points from a typical manipulation task. 
In order to minimize the number of points that are transmitted, we detect measurements on the manipulator and the close-by object by applying a combination of filters on the raw laser scans in a given scan window
extracted from the last three 2D laser scans.
First, so-called \textit{jump edges}---occurring at occluded transitions between two objects---are removed by filtering neighboring measurements by their angle.
Then, we detect edge points by applying a Sobel edge filter on the distance measurements in a scan window.
To account for edges resulting from noisy measurements, distance measurements are smoothed by a Median filter before applying the Sobel filter. 
Since dull or curvy edges may result in numerous connected edge points, we further reduce the remaining edge points by applying a line segment filter.
The line segment filter reduces a segment of connected edge points to its start and end point.
The corresponding 3D points of the remaining distance measurements are transmitted to the operator as contour points. 
Selecting contour points by filtering the distance measurements of the raw laser scans---contrary to the detection in 3D point clouds---results in a robust and efficient detector which allows us to transmit live sensory feedback over the low-bandwidth channel.

Telemetry from the robot base includes the current support polygon, estimated
COM position, emergency stop status, infrared distance measurement from
the hand, and the maximum servo temperature. Finally, the low-bandwidth link
also includes the measured audio amplitude of the right hand camera microphone,
which allows us to easily determine whether we succeeded in turning the drill on.

\subsubsection{High-bandwidth Burst Channel}

Since the connection between robot and field PC is always present,
irrespective of whether the DCE communication window is currently open,
we use this connection (red\,(2) in \cref{fig:commlink}) to transfer larger amounts
of data to the field PC for buffering.

During our participation in the DLR SpaceBot Cup \citep{stuckler2015nimbro}, we
developed a robust software module for communication between multiple ROS
masters over unreliable and high-latency networks. This module was
extended with additional features during the DRC and is now freely available
under BSD-3 license\footnote{\url{https://github.com/AIS-Bonn/nimbro_network}}.
It provides transport of ROS topics and services over TCP and UDP protocols.
Since it does not need any configuration/discovery handshake, it is
ideally suited for situations where the connection drops and recovers
unexpectedly. The high-bandwidth channel makes exclusive use of this
\textit{nimbro\_network} software. This made fast development possible, as
topics can be added on-the-fly in configuration files
without developing specific transport protocols. After DRC, additional
improvements to \textit{nimbro\_network} have been made, \eg adding
forward error correction for coping with large packet loss ratios.

The transmitted ROS messages are buffered on the field computer. The field
computer sends a constant 200\,MBit/s stream of the latest received ROS messages
to the operator station. This maximizes the probability of receiving complete
messages during the short high-bandwidth communication windows inside the
building.

The transferred data includes:

\begin{itemize}[noitemsep]
 \item JPEG-compressed camera images from all seven cameras on board, plus
   two high-resolution cut-outs of the overhead camera showing the hands,
 \item compressed\footnote{The point clouds were compressed using the PCL point cloud compression.}
   point cloud from the ego-centric 3D map (see \cref{sec:perception}),
 \item ROS log messages,
 \item servo diagnostics, and
 \item miscellaneous diagnostic ROS topics.
\end{itemize}

The 3D data received in the communication bursts is shown to the operators and
transformed into a fixed frame using the low-latency transform information received over the
low-bandwidth channel.

\subsubsection{High-bandwidth Direct Imagery}
\label{sec:comm:high_direct}

During the outside tasks, the high-bandwidth link is always available. This
opens the possibility of using streaming codecs for transmitting live imagery,
which is not possible in the inside mode, where communication blackouts would
corrupt the stream. Thus, an additional high-bandwidth channel using the
\textit{nimbro\_network} module carries H.264 encoded camera streams of the
main overhead camera and the right hand camera. The streams use
an increased frame rate of 5\,Hz to allow low-latency operator control.
These camera streams are used during the drive task for steering the car.
The channel is shown in brown\,(3) in \cref{fig:commlink}.

\section{Control}

The Momaro robot is challenging to control because of its hybrid
locomotion concept and the many DOFs involved. This section
describes the control strategies we developed.

\subsection{Kinematic Control}

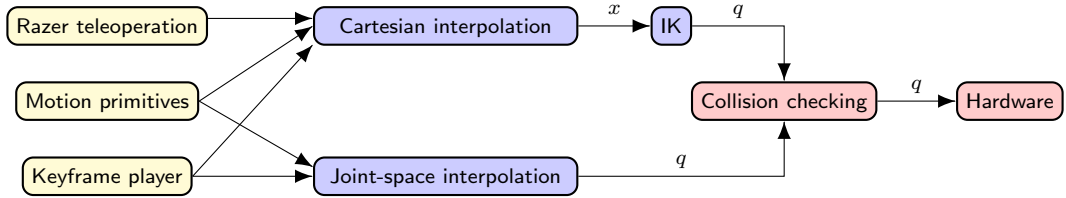
\begin{figure}
  \centering\begin{maybepreview}
\begin{tikzpicture}
    [
    font=\sffamily\footnotesize,
    n/.style={draw, thick, rounded corners,align=center,fill=yellow!20},
    int/.style={n,fill=blue!20},
    hw/.style={n,fill=red!20},
    every node/.style={text height=1.5ex,
    text depth=.25ex,
    text centered
    }
  ]

  \node[n] (keyframe) at (-1.0, -1.0) {Keyframe player};
  \node[n] (razer) at (-1.0, 1.0)       { Razer teleoperation};
  \node[n] (primi) at (-1.0, 0.0)       {Motion primitives}; 

  \node[int,minimum width=3.5cm] (joint_interpolation) at (3.5, -1.0) {Joint-space interpolation};
  \node[int,minimum width=3.5cm] (cart_interpolation) at (3.5, 1.0) {Cartesian interpolation};
  \node[int] (IK) at (6.5, 1.0) {IK};
  \node[hw] (collision) at (8.0, 0.0) {Collision checking}; 
  \node[hw] (hw) at (11.0, 0.0)  {Hardware};

 \draw[big arrow] (keyframe.east) -- (cart_interpolation.188);
 \draw[big arrow] (keyframe.east) -- (joint_interpolation.west);

 \draw[big arrow] (primi.east) -- (cart_interpolation.west);
 \draw[big arrow] (primi.east) -- (joint_interpolation.176);

  \draw[big arrow] ($(razer.east)+(0,0.1)$) -- ($(cart_interpolation.west)+(0,0.1)$);
  \draw[big arrow] (cart_interpolation) -- (IK) node [midway,above] {$x$};
  \draw[big arrow] (IK) -| (collision) node [pos=0.25,above] {$q$};
  \draw[big arrow] (collision) -- (hw) node [midway,above] {$q$};
  \draw[big arrow] (joint_interpolation.east) -| (collision) node [pos=0.25,above] {$q$};
\end{tikzpicture}
\end{maybepreview}
  \caption{Kinematic control architecture for one limb.
  The goal configuration can be specified in joint space or Cartesian space using
  the magnetic trackers, motion primitives, or the keyframe player.
  After interpolation (and IK for Cartesian poses $x$), the resulting joint
  configuration $q$ is checked for collisions and sent to the hardware.
  }
  \label{fig:kinematic_control}
\end{figure}

The kinematic control implemented in Momaro (see \cref{fig:kinematic_control})
follows a straight-forward approach. All limbs and the torso yaw joint are
considered separately.
A Cartesian or joint-space goal
configuration for a limb is defined through telemanipulation
(see \cref{sec:operator})
or dedicated motion primitives (\eg used for the DRC wall cutting task).
The Reflexxes library~\citep{kroger2011opening} is used to interpolate between
the current and desired position in Cartesian or joint space.
If concurrent limb motion is desired, Cartesian and joint-space goals can be
mixed freely for the different limbs.
The interpolation is done such that all limbs arrive at the same time.
Interpolated
Cartesian poses are converted to joint space positions via inverse kinematics.
Finally, the new robot configuration is checked for self-collisions and,
if collision-free, fed to the low-level hardware controllers for execution.

For the 7 DOF arms, we calculate the inverse
kinematics with redundancy resolution using the selectively damped least squares
(SDLS) approach~\citep{buss2005selectively}. SDLS is an iterative method based on
the singular value decomposition of the Jacobian of the current robot
configuration. It applies a damping factor for each singular value based on the
difficulty of reaching the target position. Furthermore, SDLS sets the target
position closer to the current end-effector position if the target position is
too far away from the current position. SDLS robustly computes target
position as close as possible to 6D poses if they are not within the reachable
workspace of the end-effector. Furthermore, we combine SDLS with a nullspace
optimization based on the projection of a cost function gradient to the
nullspace~\citep{liegeois1977automatic}. The used cost function is a sum of three
different components:

\begin{enumerate}
  \item Joint angles near the limits of the respective joint are punished to avoid joint limits, if possible.
  \item The difference between the robot's last and newly calculated configuration is penalized to avoid jumps during a motion.
  \item The difference from a user-specified ``convenient'' configuration and the newly calculated configuration is punished to reward this specific arm position.
    We chose this convenient configuration to position the elbow of each arm next to the body.
\end{enumerate}

For the legs, the IK problem is solved with a custom analytical kinematics solver.
Since the legs have four DOF (excluding the wheels), the solution
is always unique as long as it exists.

Calculated joint configurations are checked for self-collisions with simplified
convex hull collision shapes using the MoveIt! library\footnote{\url{http://moveit.ros.org}}.
Motion execution is aborted before a collision occurs. The operator can then move
the robot out of the colliding state by moving in another direction.

\subsection{Omnidirectional Driving}
\label{sec:omnidrive}

The wheel positions $\w{r}^{(i)}$ relative to the trunk determine the footprint
of the robot, but also the orientation and height of the robot trunk. An operator can
manipulate the positions via a graphical user interface
(see \cref{sec:operator:locomotion}) either directly for each wheel by dragging it
around, moving all wheels together (thus moving the trunk relative to the
wheels) or rotating all wheel positions around the trunk origin (thus
controlling the trunk orientation).

An operator can control the base omnidirectional driving using a joystick, which generates a
velocity command $\w{w} = (v_x, v_y, \omega)$ with horizontal linear velocity
$\w{v}$ and rotational velocity $\omega$ around the vertical axis. The velocity
command is first transformed into the local velocity at each wheel $i$:
\begin{align}
  \begin{pmatrix} v_x^{(i)} \\ v_y^{(i)} \\ v_z^{(i)} \end{pmatrix} = \begin{pmatrix} v_x \\ v_y \\ 0 \end{pmatrix} + \begin{pmatrix} 0 \\ 0 \\ \omega \end{pmatrix} \times \w{r}^{(i)} + \dot{\w{r}}^{(i)},
\end{align}
where $\w{r}^{(i)}$ is the current position of wheel $i$ relative to the base.
The kinematic velocity component $\dot{\w{r}}^{(i)}$ allows simultaneous leg
movement while driving.
Before moving in the desired direction, the wheel pair needs to rotate to the
yaw angle $\alpha^{(i)} = \textrm{atan2}(v_y^{(i)}, v_x^{(i)})$.

After all wheels are properly rotated, each wheel moves with linear
velocity $||(v_y^{(i)}, v_x^{(i)})^T||$. While driving, the robot continuously
adjusts the orientation of the ankle, using IMU information to keep the ankle yaw
axis vertical and thus retains omnidirectional driving capability.

\subsection{Semi-autonomous Stepping}
\label{sec:control:stepping}

In teleoperated scenarios, a suitable balance between
autonomous actions conducted by the robot, and operator commands has to be
found, due to the many DOF that need to be controlled
simultaneously and due to typically limited communication bandwidth.
If the terrain is not known before the robotic mission, the motion design
approach described above is not applicable. Our system addresses these scenarios
by semi-autonomously executing weight shifting and stepping actions when
required and requested by an operator.
In order to plan footsteps, the autonomous stepping module uses the 2.5D height map generated from the 3D laser measurements, described in~\cref{sec:heightmapping}.
For details on the approach, see \citet{schwarz2016hybrid}.

\begin{figure}[t]
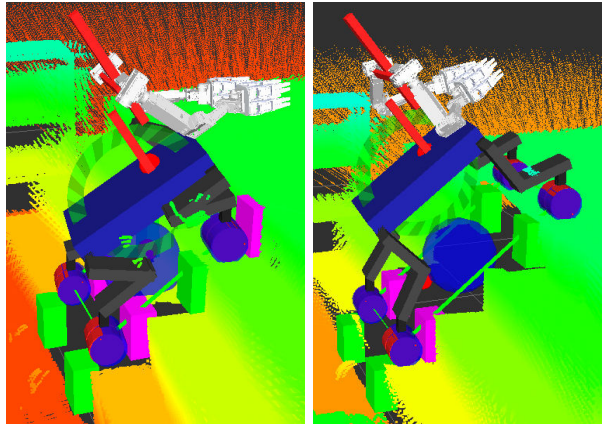

  \centering\begin{maybepreview}
  \includegraphics[height=5.6cm]{images/standing_on_stairs_crop.png}
  \includegraphics[height=5.6cm]{images/standing_on_stairs1_crop.png}\end{maybepreview}
  \vspace{-2mm}
  \caption{%
    Momaro climbing stairs in simulation. The purple and green boxes indicate
    detected obstacles which constrain the wheel motion in forward and backward
    direction, respectively.
  }
  \label{fig:sim_stairs}
\end{figure}

While the operator always retains control of the velocity of the robot base using
a joystick, steps can be triggered either automatically or manually.
The automatic mode always decides on the wheel pair which most urgently needs stepping
for continued base movement with the requested velocity. To this end, we detect
obstacles along the travel direction of the wheels (see \cref{fig:sim_stairs}).

\begin{figure}[t]
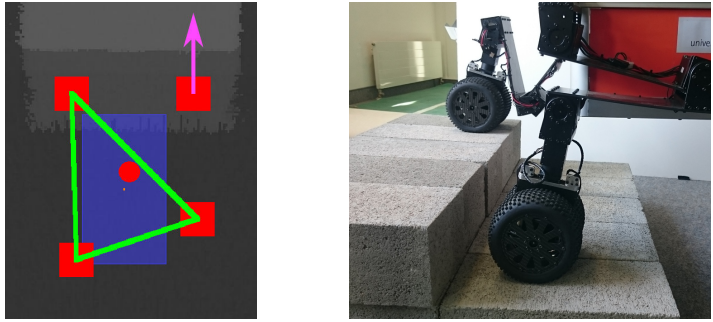

 \centering\begin{maybepreview}
 \includegraphics[height=4.2cm,clip,trim=190 170 170 100]{images/snap4_map_arrow.png}
 \hspace{1cm}
 \includegraphics[height=4.2cm]{images/DSC_0114_small.JPG}\end{maybepreview}
 \vspace{-2mm}
 \caption{
   Left: 2D heightmap of Momaro standing on two steps of a set of
   stairs in our lab.
   The robot is in stable configuration to lift the right front leg.
   Red rectangles: Wheel positions, red circle: COM, blue: robot base, green: support polygon.
   Right: The right front leg is lifted and placed on the next step.
 }
 \label{fig:stepping}
\end{figure}

To be able to lift a wheel, the robot weight must be shifted away from it.
Ideally, the 2D projection of the COM of the robot should lie
in the center of the triangle formed by the other three wheel pairs (see~\cref{fig:stepping}). This ensures
static balance of the robot while stepping. The system has three means for
achieving this goal:
\begin{enumerate}[nosep]%
  \item moving the base relative to the wheels in sagittal direction,
  \item driving the wheels on the ground relative to the base, and
  \item modifying the leg lengths (and thus the base orientation).
\end{enumerate}

All three methods have been used in the situation depicted in \cref{fig:stepping}.
The balance control behavior ensures static balance using
foot motions on the ground (constrained by the detected obstacles) and leg lengths.
If it is not possible to move the COM to a stable position, the system waits
for the operator to adjust the base position or orientation to resolve the
situation.

The stepping motion itself is a parametrized motion primitive in Cartesian space.
The target wheel position is determined in the height map as the first possible
foothold after the height difference which is currently stepped over.
As soon as the wheel is in the target position, the weight is shifted back using
balance control. The operator is then free to continue with
either base velocity commands or further steps.
\section{Operator Interface}
\label{sec:operator}

During DRC runs, we split all operation between the ``lower body operator'', and the ``upper body operator'', and a
total of
seven support operators. One support operator assists the upper body operator
by modifying his view. Two operators are responsible for managing the local
multiresolution map by clearing undesirable artifacts or highlighting parts
of the map for the upper body operator.
Another support operator monitors the hardware and its temperature during the
runs. Two more operators assist the upper body operator by triggering
additional predefined parameterized motions and grasps and are able to control
the arms and grippers in joint space as well as in task space using a graphical
user interface if necessary. While the system is designed to be
controllable using a minimum of two operators (the lower- and upper-body
operators), the actual number of operators is flexible.

\subsection{Situational Awareness}

\begin{figure}\begin{maybepreview}
\begin{tikzpicture}
    \node[anchor=south west,inner sep=0] (image) at (0,0) {\includegraphics[width=\linewidth]{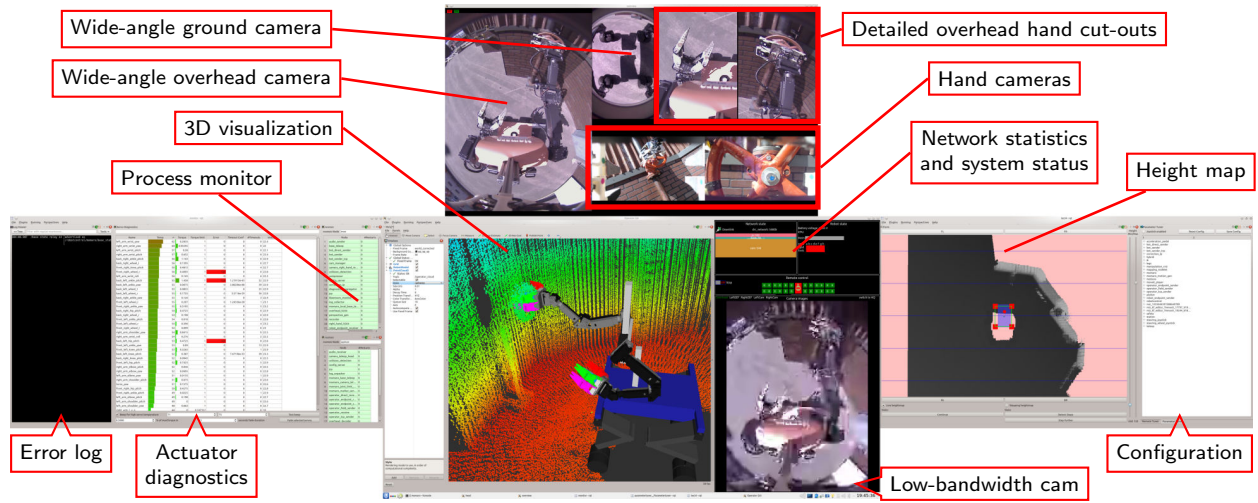}};
    \begin{scope}[
            x={(image.south east)},
            y={(image.north west)},
            font=\footnotesize\sffamily,
            line/.style={red, line width=1pt},
            every node/.style={align=center},
            box/.style={rectangle,draw=red,inner sep=0.3333em,thick}
          ]
      \node[box,anchor=north west, rectangle callout, callout relative pointer={(0.0,0.04)}] at (0.0, 0.13) {Error log};
      \node[box,anchor=north, rectangle callout, callout relative pointer={(0.0,0.04)}] at (0.15, 0.13) {Actuator\\diagnostics};
      \node[box] (3d) at (0.2,0.75) {3D visualization};
      \node[box,anchor=north] (heightmap) at (0.95,0.7) {Height map};
      \draw[line] (heightmap) -- (0.8, 0.5);

      \node[box,anchor=north east, rectangle callout, callout relative pointer={(0.0,0.04)}] at (1.0,0.13) {Configuration};
      \node[box,anchor=west,rectangle callout, callout relative pointer={(-0.02,0.0)}] at (0.7,0.03) {Low-bandwidth cam};

      \node[box] (overhead) at (0.15, 0.85) {Wide-angle overhead camera};
      \node[box] (ground) at (0.15, 0.95) {Wide-angle ground camera};

      \node[box] (overhand) at (0.8,0.95) {Detailed overhead hand cut-outs};
      \node[rectangle, draw=red, inner sep=0, fit={(0.52,0.763) (0.65,0.99)}, line width=2pt] (overhand_box) {};
      \draw[line] (overhand.west) -- (overhand_box);

      \node[box] (hand) at (0.8, 0.85) {Hand cameras};
      \node[rectangle, draw=red, inner sep=0, fit={(0.465,0.59) (0.65,0.75)}, line width=2pt] (hand_box) {};
      \draw[line] (hand.west) -- (hand_box.east);

      \node[box] (network) at (0.8, 0.7) {Network statistics\\and system status};
      \draw[line] (network.west) -- (0.63, 0.5);

      \node[box] (rosmon) at (0.15, 0.65) {Process monitor};
      \draw[line] (rosmon) -- (0.28, 0.4);

      \draw[line] (3d.east) -- (0.4,0.5);
      \draw[line] (overhead.east) -- (0.4, 0.8);
      \draw[line] (ground.east) -- (0.5, 0.9);

    \end{scope}
 \end{tikzpicture} %
\end{maybepreview}
  \caption{GUI on the main operator station, during the DRC valve task.}
  \label{fig:operator:gui}
\end{figure}

\begin{figure}[t]
  \centering\begin{maybepreview}
    \includegraphics[width=0.9\linewidth]{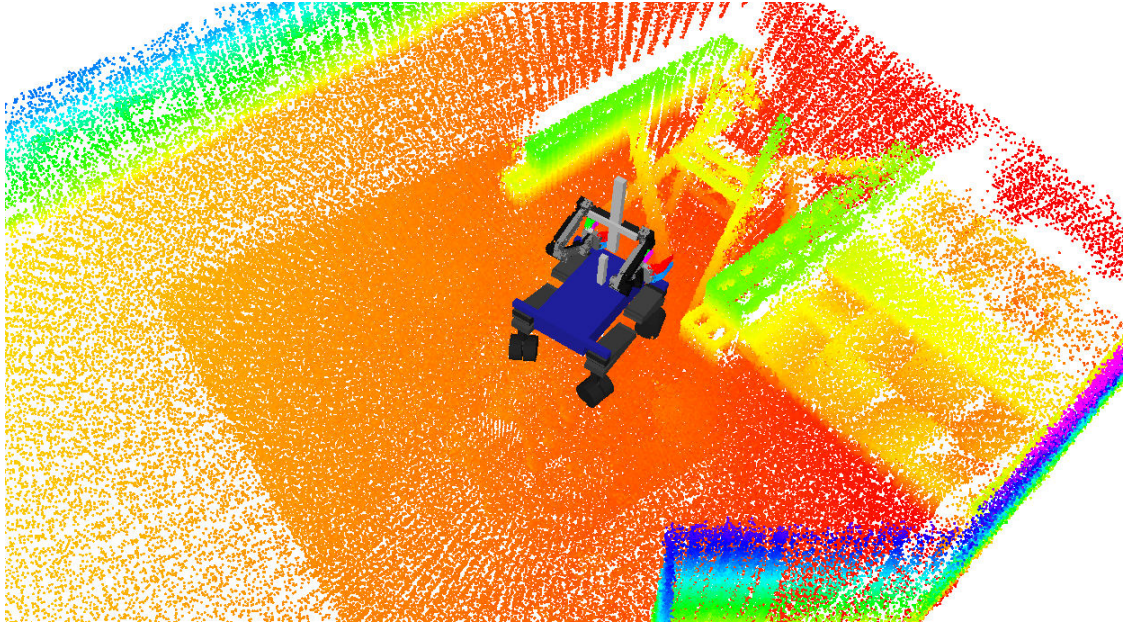}\end{maybepreview}
  \caption{Point cloud of the egocentric multiresolutional surfel map, as viewed by
    the robot operator during the debris task of the first DRC competition run.
    Color encodes height.
  }
  \label{fig:surfelmap}
\end{figure}

The main operator interface shown on the dedicated operator station computer
over four screens can be seen in \cref{fig:operator:gui}.
Operator situational awareness is gained through 3D environment visualization
and transmitted camera images.
The upper screen shows camera images from the overhead camera, ground camera,
and hand cameras. It also shows higher-resolution cut-outs from the overhead
camera centered on the hands. For all camera images, the view always shows the
last received image, independent of the image source (low-bandwidth or
high-bandwidth burst). This ensures that operators always use newest
available data.

The lower middle screen shows a 3D visualization of the robot in its environment.
Serving as main environmental representation (see~\cref{fig:surfelmap}), a downsampled and clipped map---generated from robot's egocentric map described in \cref{sec:perception}---is transmitted over the communication link.
The screen also shows the currently selected low-bandwidth image channel.

The left screen shows diagnostic information, including the ROS log (transmitted
in the bursts), actuator temperatures, torques, and errors, and process status
for nodes running on the robot and the operator station. The process monitoring
is handled by the \textit{rosmon} software\footnote{\url{https://github.com/xqms/rosmon}}.
The right screen
shows a 2D height map of the environment and allows configuration of all
system modules through a hierarchical GUI.

The support operators use notebooks connected to the operator station over Ethernet.
Using the flexibility of ROS visualization tools, the notebooks offer views customized
for the individual operator task.

\subsection{Motion Design}
\label{sec:motion}

To support fast and flexible creation of motions by a human designer, we
developed a set of motion editing tools. Motions are initially specified using
a keyframe editor. At runtime, motions can be loaded, modified to
fit the current situation, and finally executed by a player component.

\begin{figure}
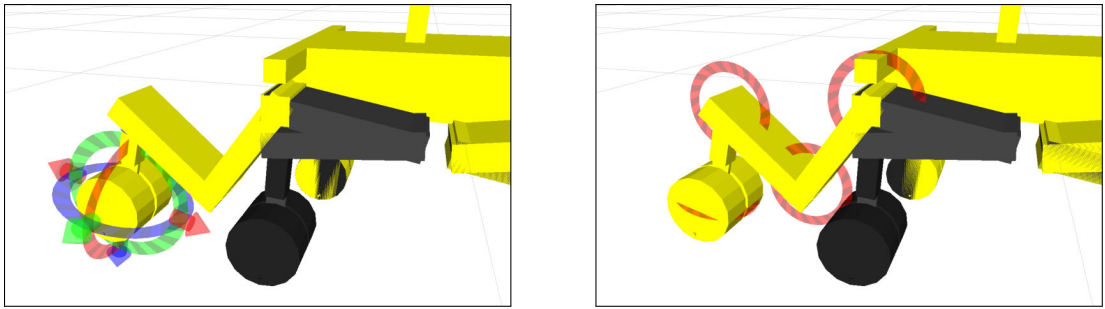

  \centering\begin{maybepreview}
  \setlength{\fboxsep}{0cm}
  \fbox{\includegraphics[clip,trim=0cm 5cm 0cm 0cm,height=4cm]{images/keyframe_ik.png}}%
  \hspace{1cm}
  \fbox{\includegraphics[clip,trim=0cm 5cm 0cm 0cm,height=4cm]{images/keyframe_jointspace.png}}\end{maybepreview}
  \caption{%
    Graphical user interface for keyframe editing. The user specifies a
    Cartesian target pose (left) or a target configuration in joint
    space (right). The yellow robot model displays the target
    configuration while the current robot configuration is shown in black.
  }
  \label{fig:keyframe_editor}
\end{figure}

The keyframe editor (see~\cref{fig:keyframe_editor}) is based on the standard
ROS RViz graphical user interface.
It shows the current robot state and the keyframe goal configuration
as 3D models. Since the robot has a large number of independent endeffectors
and internal joints, keyframes consist of multiple joint group states. For each
joint group (\eg the right arm), the user can specify either a target
configuration in joint space, or a target endeffector pose in Cartesian space.
Interpolation between the keyframes is controlled by specifying velocity
constraints. Furthermore, the user can also control the amount of
torque allowed in the motor controllers. Finally, the user can attach so-called
frame tags to the keyframe, which trigger custom behavior, such as the wheel
rolling with the motion of the leg. The tagging method allows the keyframe
system to stay isolated from highly robot-specific behavior.

The described motion design method can be used offline to pre-design fixed
motions, but it can also be used online to teleoperate the robot. In this case,
the operator designs single-keyframe motions consisting of one goal
configuration, which are then executed by the robot.
The 3D map visualization can be displayed in the
keyframe editor, so that the operator can see the current and target
state of the robot in the perceived environment.

\subsection{Immersive Bimanual Telemanipulation}
\label{sec:operator:razer}

\begin{figure}[t]
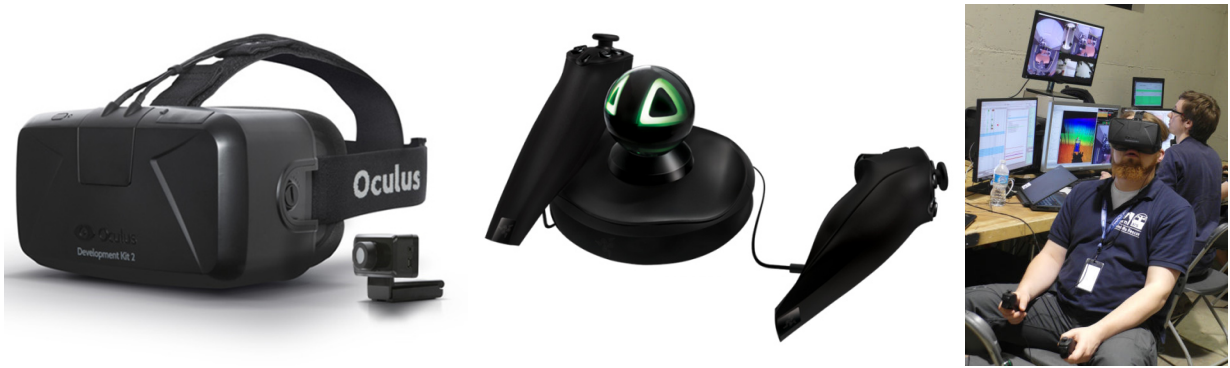

  \centering\begin{maybepreview}
  \includegraphics[height=4.8cm,clip,trim=50 0 0 0]{images/Oculus_Rift.jpg}
  \includegraphics[height=4.8cm]{images/Razer_Hydra.jpg}
  \includegraphics[height=4.8cm]{images/DRC_Finals_NimbRo_Rescue_Rehearsal_Telemanipulation.jpg}\end{maybepreview}
  \caption{Immersive telemanipulation. Left: Oculus Rift DK2 HMD. Center: Razer Hydra magnetic trackers. Right: Upper body operator using the HMD and trackers during DRC.}
  \label{fig:operator:razer}
\end{figure}

\begin{figure}
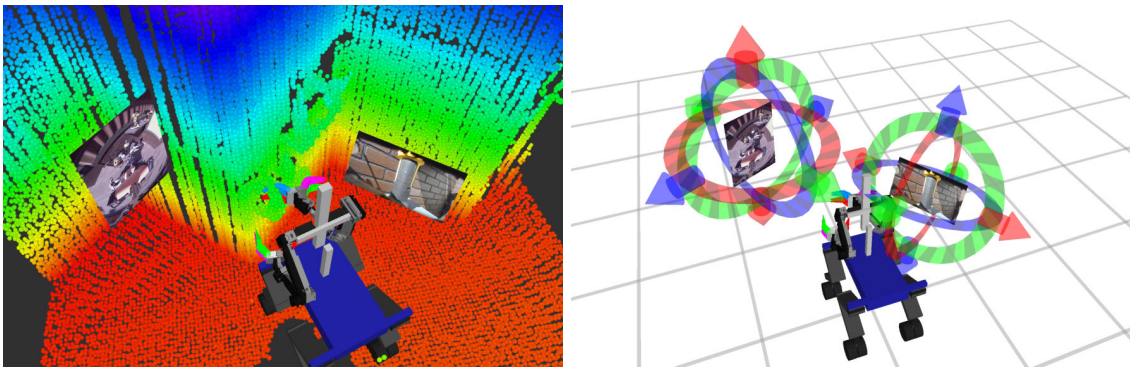

  \centering\begin{maybepreview}
  \includegraphics[height=4.8cm]{images/rviz_upper_body_operator_cutting.png}
  \includegraphics[height=4.8cm]{images/rviz_support_operator_cutting.png}\end{maybepreview}
  \caption{Left: Third person view of the upper body operator display. Right: Same scene as seen by a support operator.}
  \label{fig:hmd_and_support_view}
\end{figure}

For intuitive and flexible manipulation, a designated upper body operator is responsible for controlling the robot, using two Razer Hydra\footnote{http://sixense.com/razerhydra} controllers (see \cref{fig:operator:razer}).
To give the operator an immersive feeling of beeing inside robot in its environment, he is
wearing an Oculus Rift\footnote{https://www3.oculus.com/en-us/rift/} which
displays an egocentric view from the perspective of the robot which is based on
the generated local multiresolution map. The Oculus Rift is an HMD which
displays stereoscopic images and tracks the movement of the operator head in
6\,DOF. It uses a combination of gyroscopes and acceleration sensors to
estimate the rotation of the head and an additional camera-based tracking unit
to determine the head position. The tracked head movements of the operator are
used to update the stereoscopic view and allow the operator to freely look
around in the current scene. In addition, transferred 2D camera images can
be displayed in the view of the upper body operator to give him additional clues,
as can be seen in the left part of \cref{fig:hmd_and_support_view}.
The selection and positioning of these views
are performed by an additional support operator using a custom GUI
(see \reffig{fig:hmd_and_support_view}).

The Razer Hydra hand-held controllers (see \cref{fig:operator:razer})
use a weak magnetic field to sense the 6D
position and orientation of the hands of the operator with an accuracy of 1\,mm
and 1$^{\circ}$. The controllers have several buttons, an analog stick and a
trigger. These controls map to different actions which the upper body operator
can perform. The measured position and orientation of the operator hands are
mapped to the position and orientation of the respective robot gripper to allow
the operator to intuitively control them.  We do not aim for a one-to-one
mapping between the workspace of the robot and the reachable space of the
magnetic trackers. Instead, differential commands are sent to the robot:
The operator has to hold the trigger on the right or the left controller if
he wants to control the respective arm. Vice versa, the
operator needs to release the trigger to give up the control. This indexing
technique enables the operator to move the robot grippers to the boundaries of
the workspace in a comfortable way. Due to the limitation of the bandwidth, we
send the desired 6D poses of the end-effectors with a limited rate of
5\,Hz to the robot.

For small-scale manipulation, the operator can switch to a precision mode.
Here, motion is scaled down,
such that large movements of the controllers result in smaller movements
of the robot arms, thus enabling the operator to perform tasks with higher
accuracy. The operator also has the ability to rotate the torso
around the yaw axis using the analog stick on the left hand-held controller. The
upper body operator can trigger basic torque-based open/close gripper motionts with a button push.
More complex grasps are configured by a support operator.

In addition, the upper body operator has the ability to move the point of view
freely in the horizontal plane out of the egocentric view using the analog stick of the
right Razer Hydra controller and can also flip the perspective by
180$^{\circ}$ at the push of a button. Both features allow the operator to inspect the
current scene from another perspective.

The control system checks for self-collisions and displays the links which are
nearly in collision color-coded to the operators. The system stops the execution
of motion commands if the operator moves the robot further into nearly
self-collision. We do not check collisions with the environment, as they are
necessary to perform manipulation tasks.

\begin{figure}
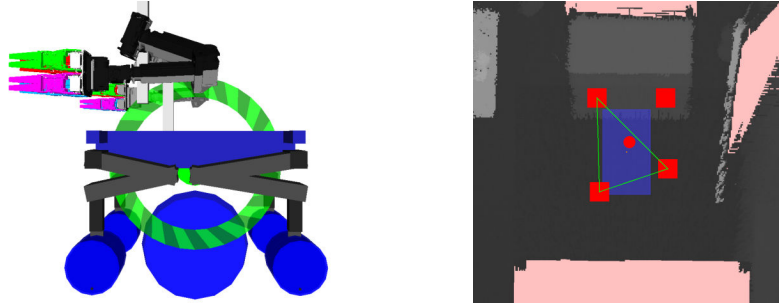

  \centering\begin{maybepreview}
  \includegraphics[height=4.0cm]{images/attitude_control.png}
  \hspace{1cm}
  \includegraphics[height=4.0cm]{images/snap4_map.png}\end{maybepreview}
  \caption{Base control GUIs.
    Left: GUI for footprint and attitude control.
    The small blue wheels can be dragged with the mouse to adjust wheel
    positions. The blue sphere controls all wheels at once, and the green ring
    can be used to modify the pitch angle of the base.
    Right: 2D heightmap of the environment. The robot base is shown in blue,
    wheels are red rectangles, the COM is a red circle. The current
    support polygon is shown in green.
  }
  \label{fig:operator:base_control}
\end{figure}

\subsection{Locomotion}
\label{sec:operator:locomotion}

During driving locomotion, the base velocity is controlled using a 4-axis
joystick. The velocity components $v_x$, $v_y$, and $\omega$ are mapped to the three corresponding
joystick axes, while the joystick throttle jointly scales all three components.
The operator can control the footprint and base attitude using a custom
base control GUI (see \cref{fig:operator:base_control}).
The operator interface for semi-autonomous stepping (see \cref{sec:control:stepping})
consists of a 2D heightmap (see \cref{fig:operator:base_control}) showing the robot
footprint, COM, support polygon and candidate step locations.

\subsection{Teleoperated Car Driving}
\label{sec:operator:car_driving}

\begin{figure}
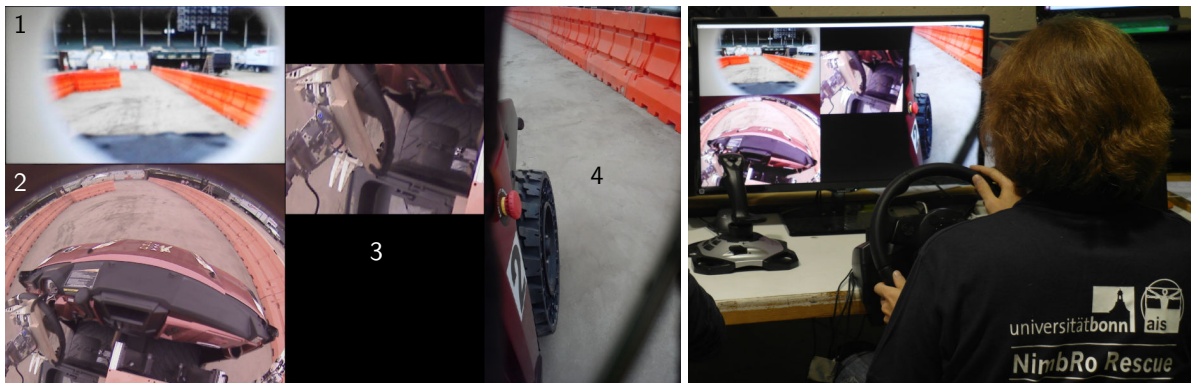

 \centering\begin{maybepreview}
 \begin{tikzpicture}
    \node[anchor=south west,inner sep=0] (image) at (0,0) {\includegraphics[height=5cm]{images/snapshot_driving.png}};
    \begin{scope}[x={(image.south east)},y={(image.north west)},font=\sffamily]
      \node[black,anchor=north west] at (0.0, 1.0) {1};
      \node[white,anchor=north west] at (0.0, 0.58) {2};
      \node[white] at (0.55, 0.35) {3};
      \node[black,anchor=east] at (0.9, 0.55) {4};
    \end{scope}
 \end{tikzpicture}
 \includegraphics[height=5cm]{images/DRC_Finals_NimbRo_Rescue_Rehearsal_Car.jpg}\end{maybepreview}
 \caption{User interface for the car driving task.
  Left: Camera view showing the center sensor head camera (1), the wide-angle overview camera (2), detail on the hand and gas pedal (3) and the right hand camera (4).
  Right: Operator using steering wheel and gas pedal during the driving task at DRC Finals.
 }
 \label{fig:operator:car}
\end{figure}

For the DRC car task (see \cref{sec:eval:drc:car}), we designed a custom operator interface
(see \cref{fig:operator:car}) consisting of a
special GUI and commercial gaming hardware controls (steering wheel and gas pedal).
The steering wheel was mapped 1:1 to the rotation of the robot hand at
the car steering wheel, while the pedal directly controlled the extension
of the front right robot leg, which pressed down on the car gas pedal.
During the driving, the responsible operator at the steering wheel uses
high-resolution imagery (see \cref{sec:comm:high_direct}) to keep track of
the vehicle and the surrounding obstacles. While sitting in
the car, the robot extends its right arm so that the operator is able
to see the right front wheel and obstacles close to the car through the
hand camera (see \cref{fig:operator:car}, image 4).

\section{Evaluation}
\label{sec:eval}

The described system has been evaluated in several simulations and lab
experiments as well as in the DARPA Robotics Challenge (DRC) Finals in June 2015,
and during the qualification runs for the DLR SpacebotCup in September 2015 \citep{kaupischdlr}.

\begin{figure}[t]
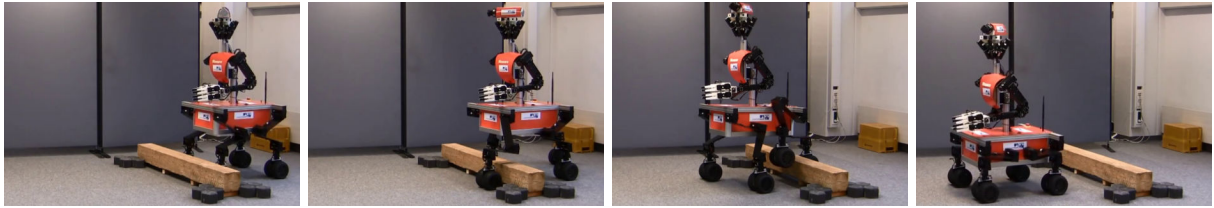

  \centering\begin{maybepreview}
  \includegraphics[height=2.7cm]{videos/bar_stepping/crop/vlcsnap-2015-09-01-14h38m57s175.png}
  \includegraphics[height=2.7cm]{videos/bar_stepping/crop/vlcsnap-2015-09-01-14h39m38s78.png}
  \includegraphics[height=2.7cm]{videos/bar_stepping/crop/vlcsnap-2015-09-01-14h40m29s78.png}
  \includegraphics[height=2.7cm]{videos/bar_stepping/crop/vlcsnap-2015-09-01-14h41m05s189.png}\end{maybepreview}
  \caption{Momaro steps over a wooden bar obstacle.}
  \label{fig:eval:bar}
\end{figure}

\begin{figure}[t]
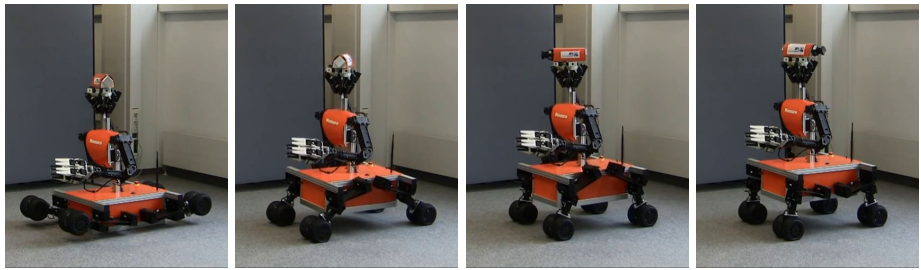

  \centering\begin{maybepreview}
  \includegraphics[height=3.5cm]{videos/stand_up/crop/vlcsnap-2015-09-15-16h05m51s72.png}
  \includegraphics[height=3.5cm]{videos/stand_up/crop/vlcsnap-2015-09-15-16h06m01s205.png}
  \includegraphics[height=3.5cm]{videos/stand_up/crop/vlcsnap-2015-09-15-16h06m06s6.png}
  \includegraphics[height=3.5cm]{videos/stand_up/crop/vlcsnap-2015-09-15-16h06m16s115.png}\end{maybepreview}
  \caption{Momaro stands up from the lowest possible configuration (base on the ground).}
  \label{fig:eval:stand_up}
\end{figure}

An early lab experiment was done to prove that Momaro fulfills the qualification
requirements of the DRC. A wooden bar obstacle (20 $\times$ 15.5 $\times$ 154\,cm) was placed in
front of the robot. For qualification, the robot was required to overcome this
obstacle. With a fixed sequence of basic stepping motion primitives,
Momaro was able to cleanly step over the obstacle (see~\cref{fig:eval:bar}).
We also showed that Momaro is capable of standing up from the lowest possible
configuration (see~\cref{fig:eval:stand_up}) to a configuration which allows 
the robot to drive, mainly using the strong hip actuators, supported by wheel 
rotation\footnote{A video of Momaro solving the qualification tasks is
available: \url{https://www.youtube.com/watch?v=PqTSPD2ftYE}}.

\subsection{DRC Finals}

The DARPA Robotics Challenge consisted of eight tasks, three of which were
mainly locomotion tasks, the other five being manipulation tasks.
Additionally, the robot had to move from one task to 
the next. Since the overall time limit for all tasks was set at one hour, quick
locomotion between the tasks was necessary for solving a large number of tasks.
Please note that the Momaro robot design was targeted for more challenging and
more numerous tasks, but DARPA lowered the number and the difficulty
of the tasks shortly before the DRC Finals.

In general, the compliance in the legs not only provided passive terrain
adaption, but also reduced the required model and kinematic precision for many
tasks by allowing the robot trunk to move compliantly in response to
environment contacts, \eg while manipulating a valve.
Furthermore, the strength of the leg actuators was also used for manipulation,
for instance when opening the door by positioning the hand under the door handle
and then raising the whole robot, thus turning the door handle upwards.

\subsubsection{Car Driving and Egress}
\label{sec:eval:drc:car}

\begin{figure*}[t]
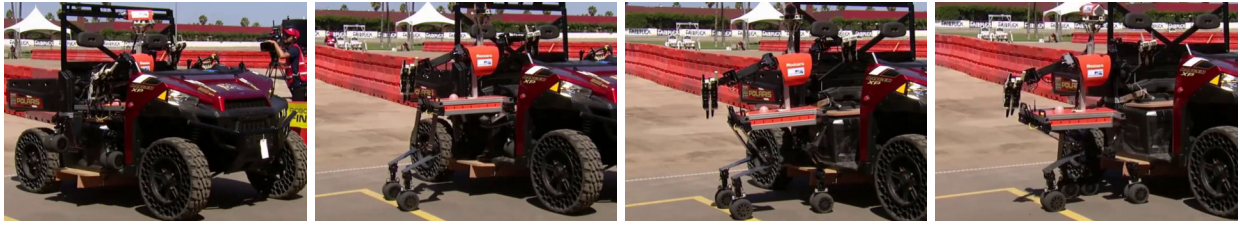

  \centering\begin{maybepreview}
  \includegraphics[height=2.9cm]{videos/car_exit/crop/vlcsnap-2015-09-01-14h54m04s0.png}
  \includegraphics[height=2.9cm]{videos/car_exit/crop/vlcsnap-2015-09-01-14h54m39s133.png}
  \includegraphics[height=2.9cm]{videos/car_exit/crop/vlcsnap-2015-09-01-14h55m07s148.png}
  \includegraphics[height=2.9cm]{videos/car_exit/crop/vlcsnap-2015-09-01-14h55m19s18.png}\end{maybepreview}
  \caption{Momaro egresses from the car at the DARPA Robotics Challenge.}
  \label{fig:eval:car}
\end{figure*}

The car task featured a Polaris RANGER XP 900 vehicle (see
\cref{fig:eval:car}), which the robot had to drive and exit. Since we did not have
access to the car before the competition, we had only a few days at the Fairplex
competition venue to determine
how to fit the robot into the car and to design an appropriate egress
motion. Even though the car task was the last we considered during
the mechanical design,
our base proved to be flexible enough to fit the robot in the car.
We extended the gas pedal with a small lever to enable Momaro to push
it with its front right leg. The steering wheel was fitted with two parallel
wooden bars to enable Momaro to turn the wheel without grasping it by
placing its fully opened gripper between the bars.
Our driving operator only had few trial runs before the actual competition. In addition,
the car engine could not be turned on during these trial runs, so the actual behavior
of the car under engine power could not be tested and trained.
Despite these limitations, we completed the car task successfully and
efficiently on the preparation day and the two competition days. We conclude that
our operator interface for driving (see \cref{sec:operator:car_driving}) is intuitive enough to allow
operation with very minimal training.
In particular, the right hand image
(see \cref{fig:operator:car}) was very helpful for keeping the appropriate
distance to the obstacles.

While many teams opted to seat their robots dangerously close to the
side of the car, so that they could exit with a single step, we placed the
robot sideways on the passenger seat and used the robot wheels to slowly drive out
of the car, stepping down onto the ground as soon as possible.
Also, some teams made extensive modification
to the car in order to ease the egressing progress, while we only added a small
wooden foothold to the car to decrease the total height which had to be overcome in
one step.
We designed an egress motion consisting of both driving and stepping components,
which made the robot climb backwards out of the passenger side of the vehicle.
Momaro successfully exited the car on the trial day and in the first
run of the competition (see~\cref{fig:eval:car}). The attempt in the second run 
failed due to an operator mistake, resulting in an abort of the egress
and subsequent reset of the robot in front of the door.

\subsubsection{Door Opening}

\begin{table}[t]
\caption{Manipulation results during the DRC Finals.}
\label{table_example}
\centering \normalsize
\begin{threeparttable}
  \begin{tabularx}{0.6\linewidth}{XXXX}
  \toprule
  \multirow{2}{*}[-0.3em]{Task} & \multirow{2}{*}[-0.3em]{Success} & \multicolumn{2}{c}{Time [min:s]} \\
  \cmidrule{3-4}
                        &                          & 1st run        & 2nd run \\
  \midrule
  Door & 2/2 & 2:25 & 0:27\\
  Valve & 2/2 & 3:13 & 3:27\\
  Cutting & 1/1 & 12:23 & -\\
  Switch & 1/1 & 4:38 & -\\
  Plug & 1/1 & - & 9:58\\
  \bottomrule
  \end{tabularx}
  \vspace{0.5em}
       \footnotesize The listed times are calculated based on a recorded video feed. All attempted manipulation tasks were successfully solved. The listed times include the time for the locomotion from the previous task to the current task.
\end{threeparttable}
\label{tab:times}
\end{table}

The first task to be completed after egressing from the vehicle is opening the door.
The door opens inwards, away from the robot. It opens either by pressing
the door handle down from above or up from below. First, the lower body operator
centers the robot manually in front of the door.
To preserve the delicate FinGripper finger tips (1.1\,mm polyamide material),
a support operator triggers a motion primitive which folds the finger tips
aside. This allows the robot to press the door handle with the 
joint servos instead. The upper body operator now uses the Razer Hydra controller to
position the left hand below the door handle. By increasing the height of the robot base
through leg extension, the door handle is pushed upwards.
Once inside the building, communication is degenerated.

After we successfully demonstrated driving the vehicle and egress from the
vehicle in our first run, we tried to open the door. On our first attempt, we
missed the door handle, as the robot was too far away from the door. After a small
robot pose correction, we succeeded. The elapsed time for this
task as well as all other attempted manipulation tasks are displayed in
Table~\ref{tab:times}.

\subsubsection{Turning a Valve}

\begin{figure*}[t!]
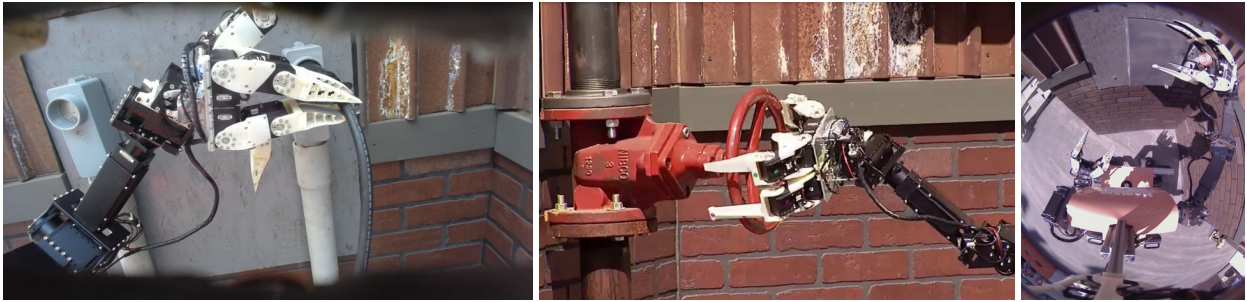
\begin{maybepreview}
  \includegraphics[height=.24\linewidth]{images/DRC_Finals_NimbRo_Rescue_Run2_Plug.jpg}
  \includegraphics[height=.24\linewidth]{images/DRC_Finals_NimbRo_Rescue_Run1_Valve_Turning.jpg}
  \includegraphics[height=.24\linewidth]{images/DRC_Finals_NimbRo_Rescue_Run1_Switch.jpg}\end{maybepreview}
  \caption{Left: Inserting the plug as seen from the right hand camera. Middle: Momaro turns the valve. Right: Flipping the switch as seen from the top-down camera. }
  \label{fig:plug_valve_switch}
\end{figure*}

This task requires the robot to open a valve by rotating it counter-clockwise by
360$^{\circ}$.
The lower body operator positions the robot
roughly in front of the valve. Then, a support operator marks the position and
orientation of the valve for the robot using an 6D interactive
marker~\citep{gossow2011interactive} in a 3D graphical user interface. After the
valve is marked, a series of parameterized motion primitives, which use the
marked position and orientation, are executed by the support operator to fulfill
the task. First, the right hand is opened widely and the right arm moves the
hand in front of the valve. The correct alignment of the hand and the valve is
verified using the camera in the right hand and the position of the hand is
corrected if the alignment is insufficient. Next, we perform the maximum clockwise 
rotation of the hand and the flexible finger tips close around the outer part of 
the valve to get a firm grasp of the valve.  Due to kinematic constraints, we can 
only turn the hand by 286$^\circ$ (5\,rad).  After that, the
hand opens again and the sequence is repeated until the valve is fully opened.
The upper body operator is not involved in this task.
We demonstrated turning the valve successfully in both runs.
During the first run, one finger tip of the right gripper slipped into the valve
and was damaged when we retracted the end-effector from the valve. We continued
the run without problems, as this was only a minor damage.

\subsubsection{Cutting Drywall}

\begin{figure}[t]
\centering\begin{maybepreview}
\begin{tabu} to 0.7\linewidth {@{}X[3]@{}X[0.1]@{}X[3.95]@{}}
  \includegraphics[width=\linewidth,render=false]{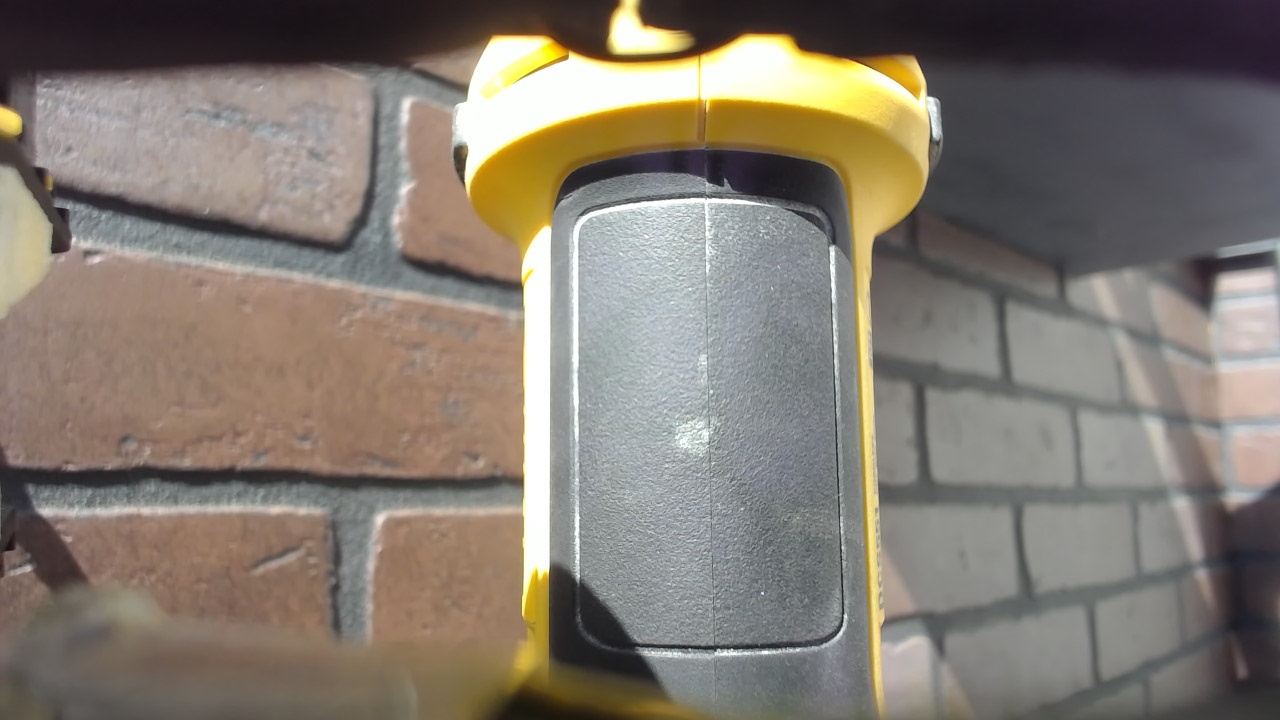} & & \multirow{3}{*}{\includegraphics[width=0.98\linewidth,render=false]{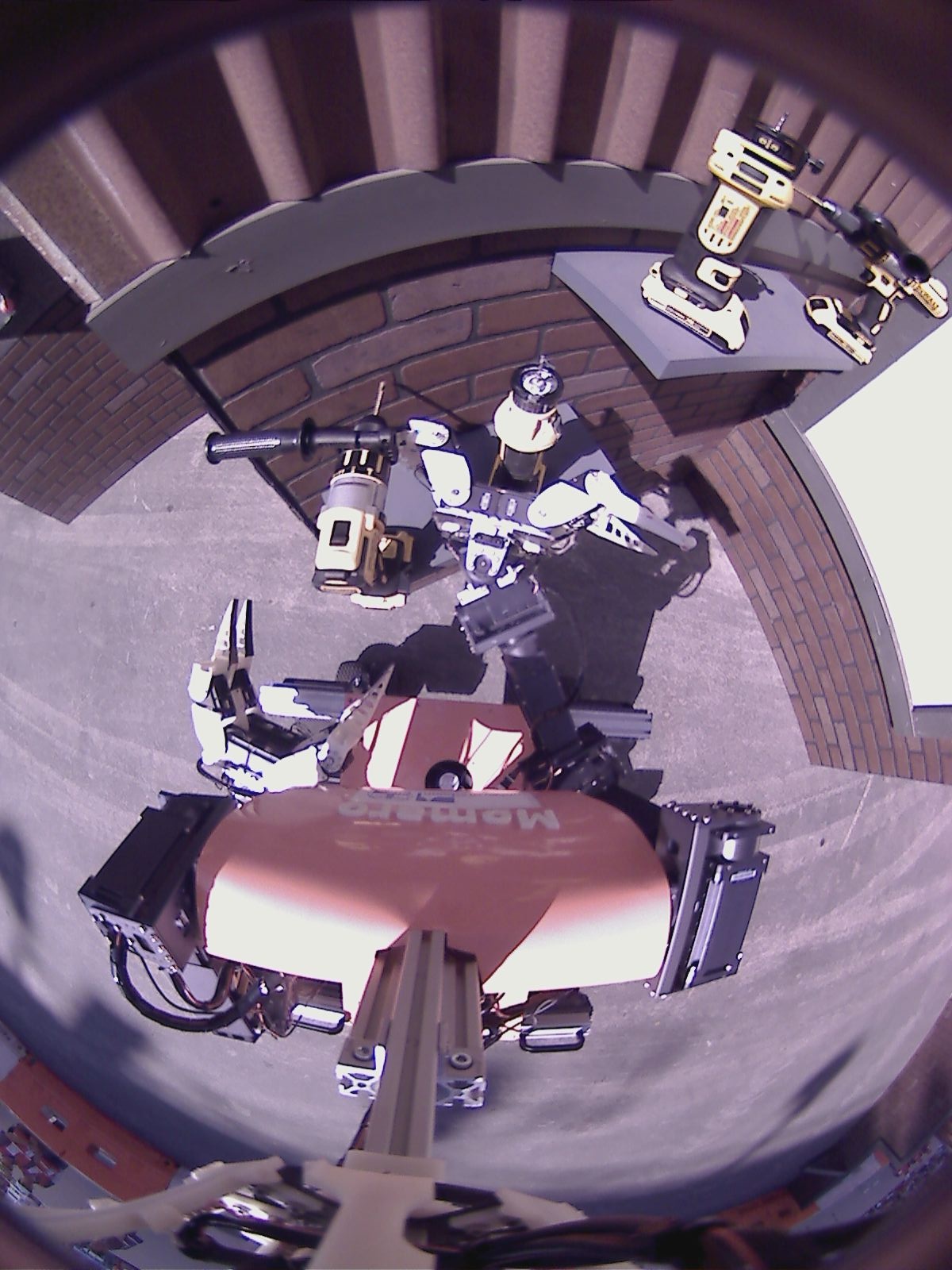}} \\
  \includegraphics[width=\linewidth,render=false]{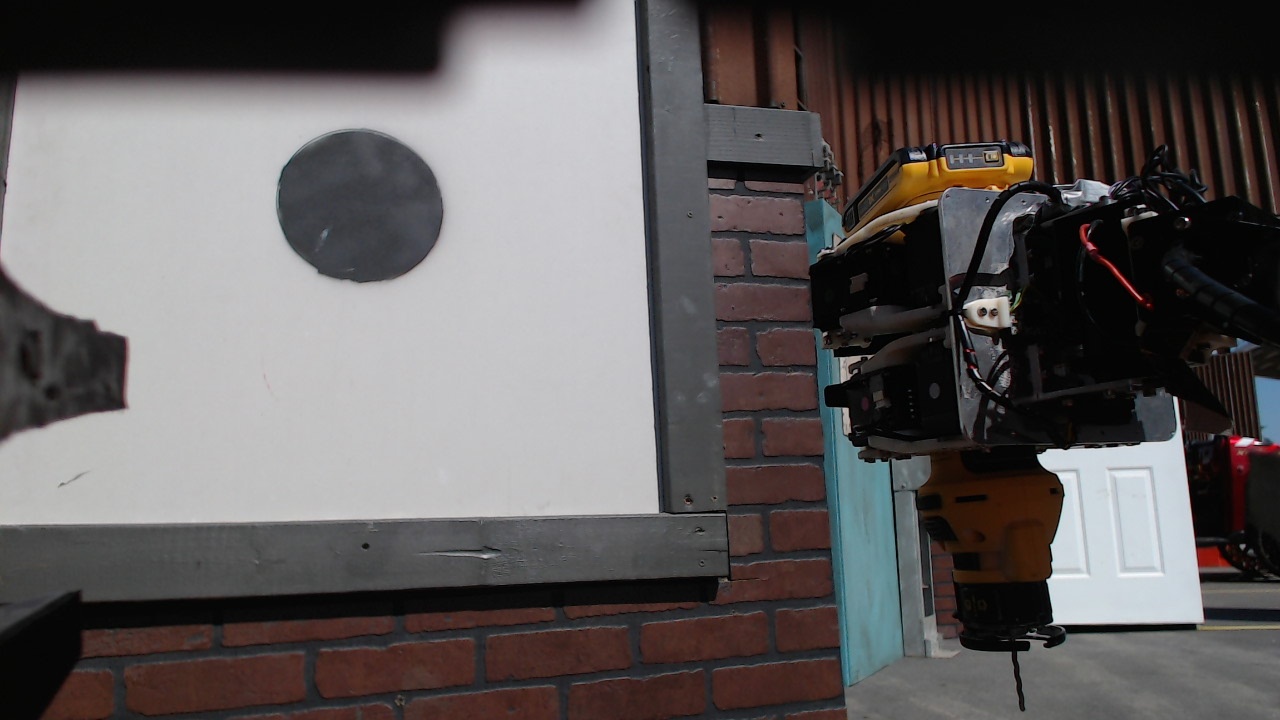} & &  \\
  \includegraphics[width=\linewidth,render=false]{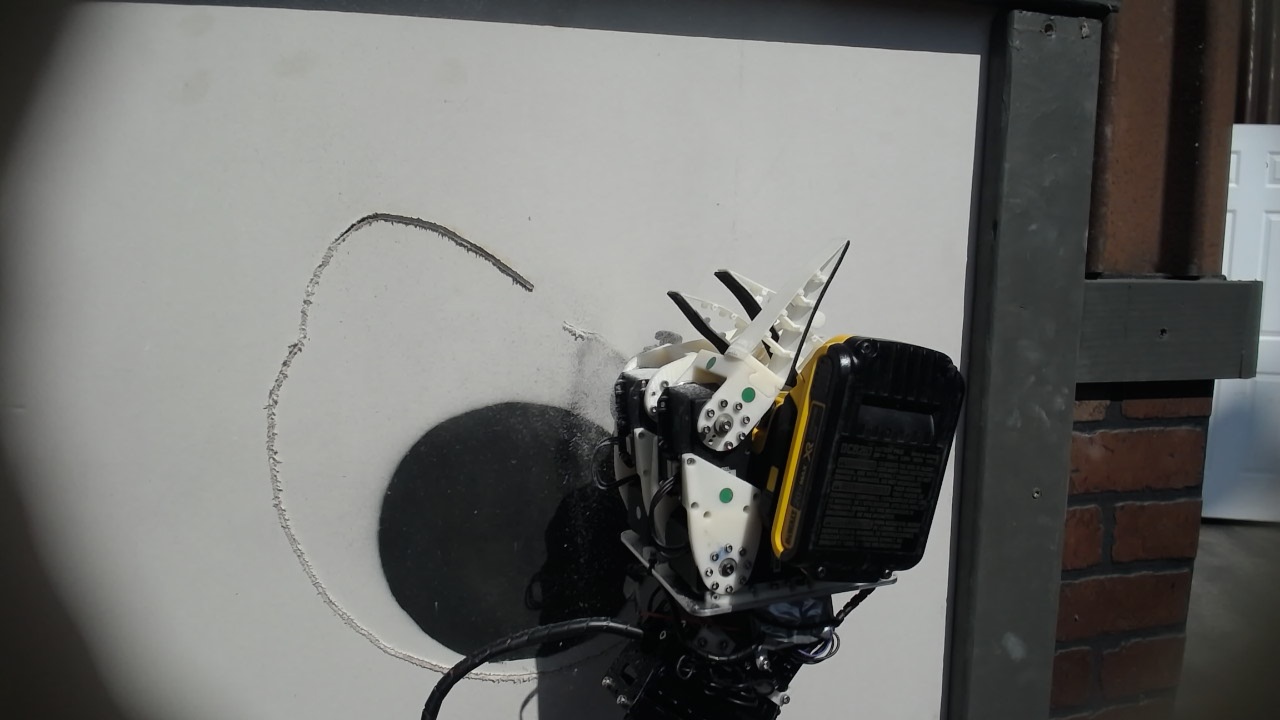} & &  \\
\end{tabu}\end{maybepreview}
  \caption{%
     Top Left: Grasping the cutting tool as seen from the right hand camera.
     Right: Same scene as seen from the top-down camera.
     Middle Left: Grasp used to switch on the tool.
     Bottom Left: Momaro cutting the drywall as seen from a sensor head camera.
  }
  \label{fig:cutting}
\end{figure}

The cutting task requires the robot to grasp one of four supplied drill
tools and use the tool to remove a marked circle from a piece of drywall by
cutting around it. We decided to use the tool which needs to be switched on only
once, instead of the tool which needs to be triggered constantly to keep
working. To switch it on, one finger of the right hand of the robot is
equipped with an additional bump.
If the tool is grasped correctly, this bump can be used to activate the trigger.
Since the tool switches off automatically after five minutes, we did not need
to design a switch-off mechanism. The tool is grasped by the
upper body operator using the Razer Hydra controller by aligning the gripper to
the tool and triggering a predefined grasp. The arm is then retracted
by the upper body operator and a support operator triggers a motion primitive
which rotates the hand by 180$^{\circ}$. As the first grasp does not close the
hand completely, the tool can now slip into the desired position. A support
operator now executes another motion to close the hand completely and switch the tool on.
After tool activation is confirmed by rising sound volume from the right hand microphone,
the upper body operator positions the tool in front of the drywall.
We fit a plane into the wall in front of the robot to automatically correct
the angular alignment. The lower body operator then drives the base forward,
monitoring the distance to the wall measured by the infrared distance sensor
in the right hand.
A parameterized motion primitive is then used to cut a circular
hole into the drywall. When the task is completed, the tool is placed on the
floor.

In our first run, we grasped the tool successfully and rotated it upside down (see
\cref{fig:cutting}). Some manual adaptation of the gripper in joint space was
necessary since the tool was initially not grasped as desired.
As we tried to cut the drywall, we became aware that the
cutting tool was not assembled correctly. Therefore, our first run was paused by
the DARPA officials and the cutting tool was replaced. The lost time was
credited to us. During our second cutting attempt, our parameterized cutting
motion primitive was not executed correctly as the robot was not properly
aligned to the drywall. Consequently, the automated cutting motion did not remove
all designated material. We noticed this error during the execution of the
motion and a support operator moved the right arm manually upwards, breaking
the drywall and fulfilling the task.

\subsubsection{Operating a Switch}

This task was the surprise task for the first run. The task is to flip a big
switch from its on-position into its off-position. After the robot was driven in
front of the switch, the upper body operator solves this task on his own. He
closes the fingers of the right hand half way using a predefined motion and then
moves the hand towards the lever of the switch. As soon as the hand encloses the
lever, the robot base is used to lower the whole robot, thus pushing the lever
down.
Since we did not have a mockup of the switch, we were not
able to train this task prior to the run. Nevertheless, we succeeded in our
first attempt.

\subsubsection{Plug Task}

This task was the surprise task for the second run. The task was to pull a plug
from a socket and insert it into a different socket which was located 0.5\,m
horizontally away from the first socket. For this task, we added additional
distal finger segments
to the left hand of the robot to increase the surface area which has
contact with the plug. During this task, a support operator controls the left
gripper using a 6D interactive marker. The interactive marker allows to move the
gripper exclusively in a fixed direction, which is difficult using the hand-held
controllers.
During the run, it took us several attempts to solve the plug
task. We used the camera in the right hand to verify that we successfully
inserted the plug into the socket as can be seen in
\cref{fig:plug_valve_switch}.

\subsubsection{Traversing Debris}

\begin{figure*}
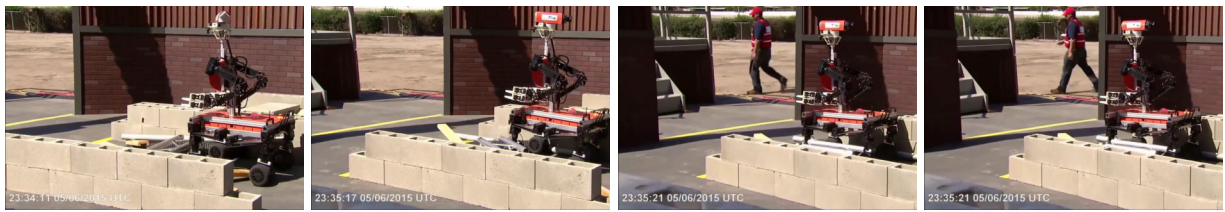

  \centering\begin{maybepreview}
  \includegraphics[height=2.7cm]{videos/rubble/crop/vlcsnap-2015-09-01-17h28m46s185.png}
  \includegraphics[height=2.7cm]{videos/rubble/crop/vlcsnap-2015-09-01-17h29m06s124.png}
  \includegraphics[height=2.7cm]{videos/rubble/crop/vlcsnap-2015-09-01-17h29m11s173.png}
  \includegraphics[height=2.7cm]{videos/rubble/crop/vlcsnap-2015-09-01-17h29m21s19.png}\end{maybepreview}
  \caption{Momaro pushes through loose debris at the DARPA Robotics Challenge.}
  \label{fig:eval:debris}
\end{figure*}

Most teams with a legged robot chose to walk over the special terrain field.
Instead, we chose to drive through the debris field using the powerful
wheels. During the trial run and first competition run, the robot simply pushed
through the loose obstacles and drove over smaller ones quite fast
(see~\cref{fig:eval:debris}). To maximize
stability, we kept the COM very low by completely folding the legs. 
During the second competition run, Momaro unfortunately got stuck
in a traverse that was part of the
debris during the second competition run. After a longer recovery procedure, the
robot still managed to solve the task, although several actuators failed due to
overheating.

\subsubsection{Stairs}

\begin{figure*}
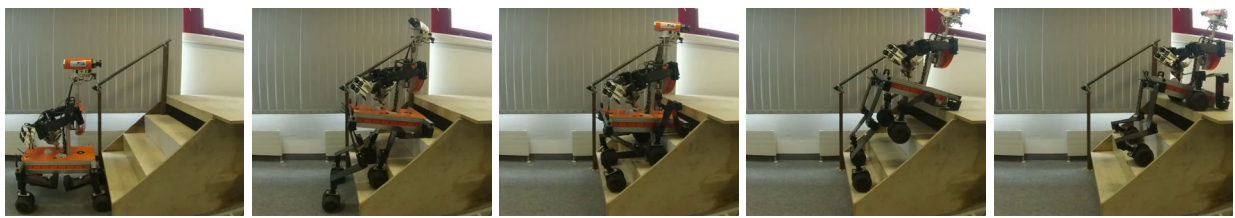

  \centering\begin{maybepreview}
  \includegraphics[height=2.75cm]{videos/stairs/crop/vlcsnap-2015-09-01-15h01m44s18.png}
  \includegraphics[height=2.75cm]{videos/stairs/crop/vlcsnap-2015-09-01-15h02m29s222.png}
  \includegraphics[height=2.75cm]{videos/stairs/crop/vlcsnap-2015-09-01-15h03m16s182.png}
  \includegraphics[height=2.75cm]{videos/stairs/crop/vlcsnap-2015-09-01-15h03m37s136.png}
  \includegraphics[height=2.75cm]{videos/stairs/crop/vlcsnap-2015-09-01-15h03m59s102.png}\end{maybepreview}
  \caption{Momaro climbs stairs, using a specially designed stair gait.}
  \label{fig:eval:stairs}
\end{figure*}

Sadly, we could not demonstrate the stairs task during the DRC Finals due to
development time constraints and the debris entanglement in the second run.
However, we were able to show that the robot is capable of climbing stairs directly
afterwards in an experiment in our lab (see~\cref{fig:eval:stairs}). To do so, 
the robot also leverages its base as a ground contact point, increasing
stability of the motion and allowing to use both forelegs simultaneously to
lift the base onto the next
step\footnote{Video: \url{https://www.youtube.com/watch?v=WzQDBRjHRH8}}.
The execution time for the experiment was only 149\,s.

\subsubsection{DRC Summary}

\begin{table}
  \centering
  \caption{Task and locomotion timings for the top five DRC teams.}
  \label{tab:toptimes}
  \begin{threeparttable}
  \begin{tabular}{lrrrrrrrrrr}
\toprule
\multirow{2}{*}[-0.3em]{Team}            & \multirow{2}{*}[-0.3em]{Day} & \multicolumn{8}{c}{DRC Tasks} & \multirow{2}{*}[-0.3em]{Locomotion} \\
\cmidrule(lr){3-10}
                &     &        Car &     Egress &       Door &      Valve &       Wall &     Rubble & Surprise\tnote{1} &     Stairs & \\
\midrule
          KAIST &   2 & \textbf{50} &          253 & \textbf{70} &    \textbf{42} & \textbf{612} &          71 &         435 & \textbf{275} &     856 \\
           IHMC &   2 &          90 &          293 &         142 &            197 &          624 &         240 &         342 &          288 &     793 \\
      Tartan R. &   1 &         117 &          836 &         524 &             55 &          813 &         111 & \textbf{72} &          298 &     491 \\
      NimbRo R. &   1 &         120 & \textbf{219} &         103 &            151 &          742 &         110 &         192 &            - &     \textbf{477} \\
     Robosimian &   1 &         266 &          518 &         126 &             86 &          798 & \textbf{54} &         311 &            - &     713 \\
\bottomrule
  \end{tabular}
  \vspace{0.5em}
  \footnotesize The better of the two runs is shown.
  The times have been roughly estimated from captured videos at DRC Finals and are by no means official.
  The ``Locomotion'' column shows the total time spent driving/walking between the tasks. Fine-alignment
  in front of a task is included. Locomotion during a task is not counted.
  \begin{tablenotes}
   \item [1] The surprise task was different between the two runs and is thus not comparable.
  \end{tablenotes}
  \end{threeparttable}
\end{table}

Our team Nimbro Rescue solved seven of the eight DRC tasks and achieved the
lowest overall time (34\,min) under all DRC teams with seven
points\footnote{Video: \url{https://www.youtube.com/watch?v=NJHSFelPsGc}}
--- the next team took 48 minutes.
This good overall result demonstrated the
usefulness of the hybrid locomotion design, the flexibility of our approach,
and its robustness in the presence of unforeseen difficulties.

As with any teleoperated system, operator training is an important aspect
of preparation, if not the most important one.
Due to the tight time schedule, our team started
testing entire runs (omitting driving, egressing and the stair task as
mentioned before) regularly about 2 weeks before the DRC finals.
Before that time, only smaller manipulation tests were possible, since the
robot hardware was not finished yet.
We feel that this short training phase was only possible due to the large
number of operators, who could independently train on their particular interface and
improve it.
This helped to identify and fix many problems in a single testing iteration.
The specialization of operators was also seen with concern, since any
operator being unavailable for any reason, \eg due to sickness, would severely limit
the crew's abilities. For this reason, we also trained backup operators for
the central tasks, in particular telemanipulation with the Oculus/Razer setup,
locomotion control, and system monitoring. Fortunately, all operators were
able to participate in the DRC Finals.

We estimated detailed task and locomotion
durations from the available video footage for the top five teams
(see~\cref{tab:toptimes}).
It is clear that these numbers are quite noisy as many factors influence the
execution time needed in the particular run. Nevertheless, some trends can be
observed. The winning team KAIST was fastest in five tasks, but took longest
for locomotion, because of transitions from standing to the kneeling
configuration. Our team NimbRo Rescue was fastest for the egress task and needed
the shortest time for locomotion, due to the fast and flexible omnidirectional
driving of our robot Momaro.

\subsection{DLR SpaceBot Cup Qualification}

\begin{figure}
  \centering\begin{maybepreview}
  \includegraphics[height=5cm]{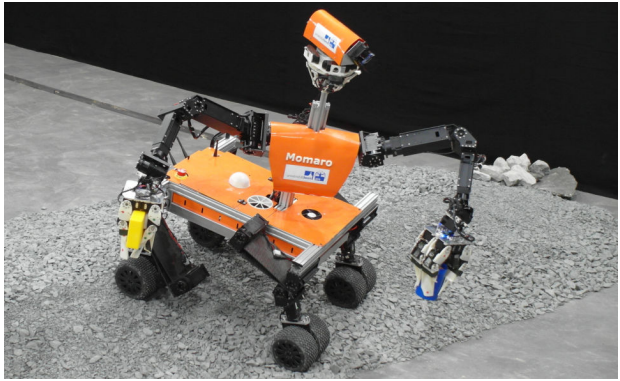}\end{maybepreview}%
  \caption{Momaro successfully participating in the DLR SpaceBot
  Cup~\citep{kaupischdlr} qualification in September 2015.}
  \label{fig:momaro_spacebot_quali}
\end{figure}

We also used Momaro to participate in the DLR SpaceBot Cup qualification runs in
September 2015 (see~\cref{fig:momaro_spacebot_quali}), where its locomotion system
allowed us to easily cross the terrain while performing manipulation tasks on
the ground with both hands. The SpaceBot Cup terrain resembles an extraterrestrial
surface and is more challenging than the smooth asphalt present at the DRC
Finals.

\subsection{Evaluation of Bimanual Telemanipulation}

\begin{figure}
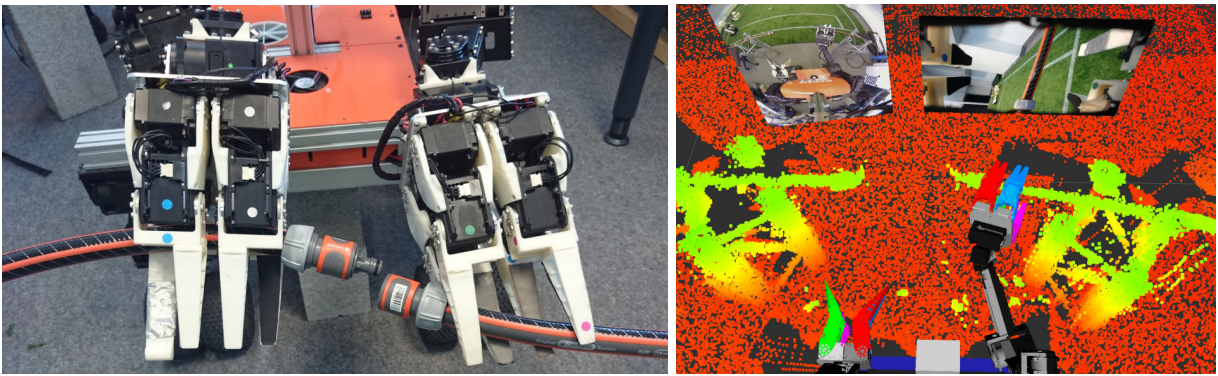

  \centering\begin{maybepreview}
  \includegraphics[height=4.9cm,clip,trim=0 0 600 0]{images/hose3.jpg}
  \includegraphics[height=4.9cm]{images/hose_screenshot.png}\end{maybepreview}
  \caption{Connecting hoses. Left: Momaro connecting two hoses. Right: Upper body operator view during the hose task. }
  \label{fig:hose}
\end{figure}

During the DRC Finals, we rarely used more than one end-effector at a time.
One example of using both hands is the plug task, where we used the right end-effector camera to 
observe the motions of the left gripper. To evaluate the bimanual teleoperation 
capabilities of our system, we designed an additional task, which exceeds the 
requirements of the DRC Finals. 

The task is to connect two flexible unmodified water hoses (see
\cref{fig:hose}). No locomotion is needed during this task, as the hoses are
placed within the reachable workspace of the robot arms. The ends of the hoses, 
which need to be connected are not placed on the floor. Instead, traverses are 
used as support for the hoses to ease grasping. This task requires bimanual 
teleoperation as the hoses are flexible and not attached to a stable base. 
Thus, the operator has to grasp both hoses with the left and right gripper,
respectively. To establish the connection between both hoses, the extension 
adapter attached to the first hose must be inserted into the connector of the 
second hose and both hoses must be pushed together in the correct angle.

One support operator assisted the trained upper body operator during the task by 
controlling the camera images which are displayed in his HMD and by triggering 
grasps. A monoscopic view from the perspective of the upper body operator can be 
seen in the right part of \cref{fig:hose}. The hoses as well as the support traverses
are clearly visible in the 3D point cloud, which provides the operator a good 
awareness of the current situation. 2D camera images are displayed to aid the 
operator with additional visual clues. Self-collision detection was switched 
off, as it might prevent close proximity of the gripper fingers, which can be 
necessary to fulfill the hose task. The operators were in a different room than 
the robot during the experiments and received information over the state of the 
robot and its environment only from the robot sensors. The communication 
bandwidth was not limited. 

We performed the hose task 11 times in our lab. The execution of one trial was 
stopped, as the upper body operator moved the right arm into the base of the 
robot as he was grasping for the right hose. The results of the remaining 10 
executions of the experiments are shown in \cref{tab:hose}. The task consists 
of three parts which are separately listed: 1. Grab the left hose with the left 
gripper, 2. Grab the right hose with the right gripper, and 3. Connect both 
hoses. On average, slightly more than three minutes were needed to complete
the whole task. The hardest part of the task was to establish the actual 
connection between both hoses, which accounted on average for more than half of 
the total elapsed time, as the upper body operator always needed more than one
attempt to connect both hoses.

\begin{table}
\caption{Execution times for the hose task (10 trials).}
\centering
\begin{tabular}{llllll}
\toprule
 \multirow{2}{*}[-0.3em]{Task} & \multicolumn{5}{c}{Time [min:s]} \\
\cmidrule(lr){2-6}
      & Avg. & Median & Min. & Max. & Std. dev. \\
\midrule
Left grasp  & 0:44 & 0:38 & 0:27 & 1:20 & 0:16 \\
Right grasp & 0:45 & 0:40 & 0:34 & 1:04 & 0:10 \\
Connect     & 1:36 & 1:32 & 1:07 & 2:04 & 0:21 \\
Total       & 3:04 & 2:57 & 2:21 & 3:51 & 0:28 \\
\bottomrule
\end{tabular}
\label{tab:hose}
\end{table}
\section{Lessons Learned}

Our participation in the DRC was extremely valuable for
identifying weak points in the described system which resulted in task or system failures.

\subsection{Mechatronic Design}

Using relatively low-cost, off-the-shelf actuators has drawbacks. In particular,
the actuators overheat easily under high torque, for example if the robot stays
too long in strenuous configurations. Such configurations include standing on the stairs with one leg
lifted. In teleoperated scenarios, delays can occur and are not easily avoided.
This hardware limitation prevented us from attempting the stairs task in our first
run, where we would have had ample time (26\,min) to move the robot up the stairs with
teleoperated motion commands. Any delay in reaching intermediate stable configurations would have
resulted in overheating and falling.
It seems that many other teams put considerable effort in active cooling of the actuators,
which would reduce this problem.
As a consequence, an improved Momaro design
would include cooled (or otherwise stronger) actuators, especially in the legs.
Also, future work will focus on further exploiting the advantages of the design by
investigating autonomous planning of hybrid driving and stepping actions, thus
allowing fluid autonomous locomotion over rough terrain and avoiding overheating
of the actuators.

Designing the legs as springs allowed us to ignore smaller obstacles and also
provides some compliance during manipulation, reducing the needed precision.
However, we also encountered problems: During our first competition run, our field
crew was worried that the robot would fall during the drill task, because one leg had
moved unintenionally far below the base, reducing the support polygon size. The deviation
was entirely caused by the springs and thus not measurable using joint encoders.
Future compliant designs will include means to measure deflection of the compliant parts,
such that autonomous behaviors and the operator crew can react to unintended configurations.

\subsection{Operator Interfaces}

In particular, operator mistakes caused failures for many teams as the reports
collected in \citep{atkesondrc} indicate.
Our second run suffered from an operator mistake (triggering the wrong motion) as well,
which could have been avoided by a better user interface.
In particular, our operator interfaces were not designed to protect against
dangerous operator commands.
In the future, we will strive to anticipate possible operator mistakes and
develop means to prevent them. Also, unanticipated situations could be
detected on the robot side, resulting in an automatic pause of the
semi-autonomous behaviors.
More time for operator training would also reduce the number of operator
mistakes.

As a second issue in our second run, the robot got entangled with a piece of
debris. Since we did not train this situation, our operator crew lost a lot of
time and power to get out of it. As a consequence, the robot actuators
overheated, which made the situation even worse.
The only possible conclusion for us is that our situational awareness was not
good enough to avoid the situation, which might be improved by mounting
additional sensors in front of the robot. Additionally, recovery from
stuck/overheated situations should be assisted by the user interface and
trained by the operator crew.

\subsection{Sensors}

It may be not a new insight, but especially our group---mainly focussing on autonomy---was surprised by the
usefulness of camera images for teleoperation. In the (autonomous) robotics community
there seems to be a focus on 3D perception, which is understandable for autonomous operation.
But color cameras have distinct
advantages over 3D sensors: They are cheap, work in harsh light conditions,
and images are easily interpretable by humans. As a result, Momaro carries seven cameras,
placed in strategic positions to be able to correctly judge situations from remote.
This strongly augments the 3D map from laser measurements.

\subsection{Preparation Time}

Our team found a viable solution for the stairs task in the night before our
second run. One could argue, that we would have needed one more day of preparation
to solve all tasks. Of course, other factors could have easily kept us from reaching this
goal as well, as seen in our second run. Nevertheless, while preparation time for
competitions is always too short, in our case it was maybe especially so.
\section{Conclusion}

In this paper, we presented the mobile manipulation robot Momaro
and its operator station
and evaluated
its performance in the DARPA Robotics Challenge, the DLR SpaceBot Cup
qualification, and several lab experiments. Novelties include a hybrid
mobile base combining wheeled and legged locomotion and our immersive approach to
intuitive bimanual manipulation under constrained communication.
The great success of the developed robotic platform and telemanipulation
interfaces at the DRC has demonstrated the feasibility, flexibility and
usefulness of the design.

To solve complex manipulation tasks, our operators currently rely on 3D point
clouds, visual and auditory feedback, and joint sensors from the robot.
Additional touch and force-torque sensing in combination with a force feedback
system for the upper body operator could potentially improve the manipulation
capabilities of the human-robot system. This could, for example, be beneficial
for peg-in-hole tasks such as the plug task during the DRC or the hose task,
which require precise and dexterous manipulation skills.

Our telemanipulation system has currently only a low degree of autonomy and
instead requires multiple human operators to control it. This allows our team to
easily react to unforeseen events. However, the number of operators needed is
quite high and so many trained operators are not always available. Therefore, it
is necessary to add more autonomous monitoring and operator assistance functions
to make the system manageable by fewer operators. Furthermore, the cognitive load on the
operators could be reduced by carrying out more tasks autonomously.

\bibliographystyle{apalike}
\bibliography{references}

\end{document}